%% file: main_arxiv.tex
\documentclass[sigplan,10pt]{acmart}
\setcopyright{none}
\renewcommand\footnotetextcopyrightpermission[1]{}
\usepackage{graphicx} 
\usepackage{xcolor}

\usepackage[most]{tcolorbox}
\usepackage{xspace}
\usepackage{textcomp}
\newcommand{\ctga}{\textsc{VeriAttn}\xspace}
\newcommand{\sect}{\textsection}
\usepackage{algorithm}
\usepackage{algpseudocode}

\newtheorem{lemma}{Lemma}
\usepackage{subcaption}
\usepackage{booktabs}
\usepackage{multirow}
\usepackage{tabularx}

\AtBeginDocument{%
  }

\begin{document}


\title{Communication-Efficient Verifiable Attention \\ for LLM Inference}


\author{Ziqun Chen}
\email{ziqun001@e.ntu.edu.sg}
\affiliation{%
  \institution{Nanyang Technological University}
  \country{Singapore}
}

\author{Ming Wu}
\email{ming@0g.ai}
\affiliation{%
  \institution{Zero Gravity Labs}
  \city{San Francisco}
  \country{USA}
}

\author{Michael Heinrich}
\email{michael@0g.ai}
\affiliation{%
  \institution{Zero Gravity Labs}
   \city{San Francisco}
  \country{USA}
}

\author{Jason Zeng}
\email{jason@0g.ai}
\affiliation{%
  \institution{Zero Gravity Labs}
    \city{San Francisco}
  \country{USA}
}

\author{Huiying Lan}
\email{huiying.lan93@gmail.com}
\affiliation{%
  \institution{Zero Gravity Labs}
    \city{San Francisco}
  \country{USA}
}

\author{Tianwei Zhang}
\email{tianwei.zhang@ntu.edu.sg}
\affiliation{%
  \institution{Nanyang Technological University}
  \country{Singapore}
}

\author{Rui Tan}
\email{tanrui@ntu.edu.sg}
\affiliation{%
  \institution{Nanyang Technological University}
  \country{Singapore}
}

\begin{abstract}
Computation integrity of remote large language model (LLM) serving can be questionable. For conventional deep neural networks (DNNs), the existing TEE-shielded DNN partitioning (TSDP) approach uses Trusted Execution Environment (TEE) to compute non-linear components and verify the integrity of linear components offloaded to an untrusted GPU.
However, directly applying TSDP to Transformer-based LLMs incurs significant TEE
computation and TEE-GPU communication overhead.
This paper presents Communication-efficient TEE-GPU Attention (\ctga) for accelerating verifiable LLM inference. 
\ctga offloads both linear and non-linear computations of attention to the GPU, while TEE performs verification. 
Moreover, for prefill, \ctga uses a two-level
pipeline to overlap data
movement, TEE pre-/post-processing, and GPU computation. 
For decoding, when the key-value cache exceeds available GPU memory,
\ctga partitions attention across TEE and GPU to reduce repeated
key-value transfers.
Evaluation on an Intel TDX platform shows that \ctga achieves 2.60-3.38$\times$ and 3.86-5.42$\times$ acceleration over TSDP for 6k-token prompts and 10k-token outputs during prefill and decoding, respectively.
\end{abstract}

\pagestyle{plain}
\maketitle







\input{Sections/Introduction}

\input{Sections/Background}

\input{Sections/Motivation}

\input{Sections/System_Design}

\input{Sections/Evaluation}

\input{Sections/Discussion}

\input{Sections/Conclusion}

\bibliographystyle{plain}
\bibliography{ref_arxiv}

\input{Sections/Appendix}

\end{document}

%% file: Sections/Introduction.tex
\section{Introduction}

Transformer-based large language models (LLMs)~\cite{vaswani2017attention} have become the
foundation of modern artificial intelligence systems and are increasingly deployed through cloud-based serving, commonly known as \emph{LLM-as-a-Service} (LLMaaS)~\cite{yu2022orca,fu2024serverlessllm,hu2025deepserve}.
However, LLMaaS has an inherent trust problem: when users submit inputs to LLMaaS, the software and
hardware stacks executing the inference may not be fully trusted by the users. In this paper, we particularly focus on the \textit{integrity} dimension of the trust, where the adversary controls privileged software or exploits system vulnerabilities to tamper with intermediate computations or manipulate model outputs. This trust issue becomes more pronounced when LLMaaS engages an {\em edge cloud} consisting of third-party edge computing powers closer to the users.

Existing approaches for verifiable inference mainly follow two directions.
First, {\em cryptography-based verification} approaches use cryptographic proof systems, such as zero-knowledge
proofs~\cite{ghodsi2017safetynets,liu2021zkcnn,sun2024zkllm,peng2025survey}, to verify that the outsourced
inference results are correctly computed.
Their main advantage is strong formal integrity guarantees.
However, they typically require encoding model execution into arithmetic circuits or constraint
systems, which incurs prohibitive proof generation overhead for large Transformer models and long sequence attention.
Second, {\em hardware-assisted} approaches use Trusted Execution Environment (TEE) as a practical hardware root of
trust to protect code and data from compromised privileged software.
Recent virtual machine (VM)-based TEEs, such as Intel TDX~\cite{aktas2023intel} and
AMD SEV~\cite{kaplan2016amd}, protect the memory and execution state of the entire VM from external.
As they provide large
protected memory spaces of up to terabytes and better software compatibility, they are suitable for hosting LLM
states.
However, TEEs still have limited computational capability for executing LLM inference compared with GPUs.

\emph{TEE-shielded DNN partitioning} (TSDP)
paradigm~\cite{tramerslalom,sun2023shadownet,zhang2024no,wang2025game} has been proposed to draw TEE's and GPU's respective advantages.
It uses TEE as the root of trust, while offloading
computation-intensive, parallelizable operators to an untrusted GPU.
For conventional deep neural networks (DNNs), such as convolutional neural networks (CNNs), TSDP
typically keeps lightweight non-linear operators in TEE, while offloading linear operators to the GPU. When the GPU returns the results, TEE performs probabilistic verification over them using Freivalds' algorithm~\cite{freivalds1977probabilistic}.
This design is effective when the intermediate activations are relatively small and the non-linear
operators such as rectified linear unit (ReLU) and pooling are lightweight.
As a result, for DNNs, TSDP significantly outperforms both fully TEE-based and cryptography-based
approaches in terms of runtime efficiency~\cite{ng2021goten,liu2021zkcnn,shen2022soter}.

Unfortunately, Transformer-based LLMs have distinct computation and communication patterns from conventional DNNs. This makes the straw man proposal of applying TSDP to LLM inefficient, in both LLM's {\em prefill} and {\em decoding} phases, as well as TEE-GPU collaboration. Three concrete challenges arise, as explained below.

First, during the prefill phase, self-attention produces large intermediate attention states, in which the
attention score and attention weight matrices scale quadratically with the prompt length.
To follow TSDP's workflow of splitting attention between GPU-side matrix multiplications and TEE-side
verification and non-linear processing, large intermediate states need to be moved across the TEE-GPU boundary, making prefill
communication-heavy for long prompts.
Moreover, the used \texttt{SoftMax} is no longer a lightweight non-linear operator to TEE, since it requires
exponentiation and normalization over large attention matrices.

Second, the decoding phase under TSDP is subjected to a tightened communication bottleneck.
Although model weights may fit into GPU memory, the key-value (KV) cache grows with the sequence length
and the number of concurrent requests, and can eventually exceed the available GPU memory. To address this issue, modern LLM serving systems offload overflowed KV cache blocks to the host memory~\cite{kwon2023efficient,lee2024infinigen,liu2025lmcache}.
To implement TSDP, 
these non-resident KV entries reside in TEE-protected host memory. As such, the GPU-centric decoding workflow must repeatedly move them across the TEE-GPU boundary for attention computation.
Since decoding attention is performance-bounded by the memory bandwidth and of low arithmetic intensity on both GPU and
TEE~\cite{he2024fastdecode,zhu2025nanoflow}, the limited computation in each decoding
step cannot amortize repeated KV cache movement.

Third,
beyond the above data volume-related challenges, TSDP exhibits low efficiency due to its serialized executions across the TEE and GPU.
During both the prefill and decoding phases, each offloaded step alternates between GPU computation, data transfer, TEE-side verification or
non-linear processing, and the subsequent GPU execution. As such, one processor often remains idle while waiting for data
transfers or for the other processor to complete computation.
This serialization reinforces the TEE-GPU communication and the TEE-side processing as performance bottlenecks.

The above observations motivate a new design to enable attention computation to be executed across the TEE and GPU in a communication-efficient manner while preserving computation integrity.
To this end, we present \ctga,
a communication-efficient TEE-GPU
attention framework for accelerating verifiable LLM inference.
\ctga addresses TSDP's major overheads for self-attention computation by two new mechanisms.

First, \ctga offloads both linear and non-linear attention computations to the GPU, while
the TEE performs only lightweight integrity verification and pre-/post-processing.
Since attention computation produces floating-point intermediate results, we devise two new approaches
to verify the integrity of matrix multiplications and
\texttt{SoftMax}'s exponentiation.
This keeps the main attention path on the GPU, allowing the GPU to proceed without waiting for
TEE-side execution and transfer.
As a result, \ctga avoids the related communication round trip and reduces both
TEE-side computation and TEE-GPU idle time while preserving integrity.

Second, we design phase-specific execution workflows for prefill and decoding.
In prefill, \ctga organizes the TEE and GPU processes into a two-level pipeline over attention
head blocks and exponentiation row tiles.
This overlaps TEE memory copies, data transfers, GPU computation, and verification inside the TEE,
thereby reducing synchronization stalls.
In decoding, \ctga uses a collaborative TEE-GPU execution strategy for attention dominated by
KV cache access.
The GPU processes the persistent GPU KV cache and selected KV blocks that are offloaded temporarily, while
the TEE locally processes the remaining non-resident KV entries.
This allows TEE-side attention to contribute to the attention computation while reducing KV transfer
overhead across the TEE-GPU boundary.

We evaluate \ctga on an Intel TDX platform over
four Transformer models
of different sizes (3/8/14B) and architectures (LLaMA/Qwen/Phi).
We compare \ctga{} with two TEE-based solutions: (i) TSDP and (ii) {\em Full-TEE}, which executes protected computation
entirely in TEE.
At a prompt length of 6,000 tokens, \ctga improves
full-model {\em time to first token} (TTFT) by $2.60$-$3.38\times$ over TSDP and $3.14$-$5.19\times$
over Full-TEE. For long-context decoding with an output length of 10,000 tokens,
\ctga improves full-model {\em time per output token} (TPOT) by $3.86$-$5.42\times$ over TSDP and
$2.21$-$3.15\times$ over Full-TEE. We also comparatively benchmark \ctga and zkLLM~\cite{sun2024zkllm} that uses cryptographic proof generation for
verifiable LLM inference.
Although zkLLM provides cryptographic soundness with negligible soundness error~\cite{sun2024zkllm}, 
it suffers two orders-of-magnitude longer TTFT and TPOT.

{\em Paper organization}:
\sect\ref{sec:background} presents background and reviews related work.
\sect\ref{sec:motivation} presents motivating benchmark results.
\sect\ref{sec:system_design} presents the design of CTGA.
\sect\ref{sec:eval} presents evaluation results.
\sect\ref{sec:discuss} discusses several relevant issues.
\sect\ref{sec:conclude} concludes this paper.

%% file: Sections/Background.tex
\section{Background and Related Work}
\label{sec:background}

\subsection{Transformer Architecture}
Transformer-based LLMs~\cite{vaswani2017attention} consist of $N$ stacked Transformer blocks, each containing a multi-head self-attention module with $H$ attention heads and a feed-forward layer. 
The self-attention module is central to Transformer, allowing each token to
aggregate contextual information from other positions in the sequence.
LLM inference consists of \emph{prefill} and \emph{decoding} phases.

Consider one attention head, during prefill, the model processes all prompt tokens in parallel.
For an input sequence of $L$ tokens, the hidden states are projected into
query, key, and value matrices
$\mathbf{Q},\mathbf{K},\mathbf{V}\in\mathbb{R}^{L\times d_h}$,
where $d_h$ denotes the per-head hidden dimension.
Self-attention first computes the scaled and masked attention score matrix
$\mathbf{S}=\mathbf{Q}\mathbf{K}^{\top} / \sqrt{d_h}+\mathbf{M}$
where $\mathbf{M}$ is the causal mask that assigns $-\infty$ to future-token
positions, ensuring that each token can attend only to itself and preceding
tokens. 
We define the attention states for accumulation and normalization in attention as follows.
\begin{equation}
    \mathbf{E}=\exp(\mathbf{S}),\qquad
    \mathbf{Z}=\mathbf{E}\mathbf{1},\qquad
    \mathbf{U}=\mathbf{E}\mathbf{V},
    \label{eq:attn-unnormalized}
\end{equation}
where $\mathbf{E}$ is the unnormalized attention-weight matrix,
$\mathbf{Z}$ is the row-sum accumulation term, $\mathbf{U}$ is the
unnormalized value aggregation and $\mathbf{1}$ is the all-one vector. 
The final normalization produces the attention output:
\begin{equation}
    \mathbf{O}
    =
    \mathbf{U}\oslash\mathbf{Z}
    =
    \texttt{SoftMax}(\mathbf{S})\mathbf{V},
    \label{eq:attention}
\end{equation}
where $\oslash$ denotes row-wise division with $\mathbf{Z}$ broadcast along the
hidden dimension. 
Since prefill performs attention over the full prompt, it involves large matrix operations such as
$\mathbf{Q}\mathbf{K}^\top$ and generates attention states whose size scales quadratically with the sequence length $L$.
The key and value representations of the prompt tokens are stored as the key-value (KV) cache for
later decoding.

During decoding, tokens are generated autoregressively, one at a time.
At each decoding step $t$, where $t$ starts from $L+1$, the model computes
$\mathbf{q}_t,\mathbf{k}_t,\mathbf{v}_t\in\mathbb{R}^{d_h}$ for the new token,
appends $\mathbf{k}_t$ and $\mathbf{v}_t$ to the KV cache, and uses
$\mathbf{q}_t$ to attend over all cached keys and values. Let
\[
    \mathbf{K}_{\le t}
    =
    [\mathbf{k}_1^\top;\ldots;\mathbf{k}_t^\top]
    \in\mathbb{R}^{t\times d_h},
    \quad
    \mathbf{V}_{\le t}
    =
    [\mathbf{v}_1^\top;\ldots;\mathbf{v}_t^\top]
    \in\mathbb{R}^{t\times d_h}.
\]
The decoding attention states and attention output are
\begin{equation}
    \mathbf{e}_t
    =
    \exp\left(
        \frac{\mathbf{q}_t^\top\mathbf{K}_{\le t}^{\top}}{\sqrt{d_h}}
    \right),
    z_t=\mathbf{e}_t\mathbf{1},
    \mathbf{u}_t=\mathbf{e}_t\mathbf{V}_{\le t},
    \quad
    \mathbf{o}_t=\frac{\mathbf{u}_t}{z_t}.
    \label{eq:attention-decoding}
\end{equation}
Here $\mathbf{e}_t$, $z_t$, and $\mathbf{u}_t$ are the decoding counterparts of
$\mathbf{E}$, $\mathbf{Z}$, and $\mathbf{U}$ for the current attention row.
Although each decoding step generates only one token, 
it must repeatedly access the accumulated KV cache.
As the generated sequence grows, the KV cache increases linearly with the decoding length, 
making
KV-cache storage and movement a critical bottleneck in long-context LLM serving.
Overall, long-sequence inference places greater pressure on both attention computation and KV-cache access, which calls for optimizations to reduce computation overhead and data movement.

\subsection{Trusted Execution Environments (TEEs)}
\label{subsec:tee_background}

TEEs provide
hardware-enforced isolation for security-critical computation.
A TEE protects the confidentiality and integrity of code and data inside the protected domain, 
even when privileged software (e.g., operating system, hypervisor, or device drivers) is compromised.
Early enclave-based TEEs, such as Intel SGX~\cite{costan2016intel}, are constrained by the limited
capacity of the Enclave Page Cache (EPC).
Exceeding the EPC capacity triggers encrypted paging and incurs substantial performance degradation.
Recent VM-based TEEs, such as Intel TDX~\cite{aktas2023intel} and
AMD SEV~\cite{kaplan2016amd}, protect the memory and execution state of an entire VM
from the untrustworthy hypervisor or host OS.
Their larger protected memory space and better software compatibility make them more suitable 
for hosting large model states and a more practical substrate for cloud LLM inference.

When a VM-based TEE interacts with external devices, it typically distinguishes between
\emph{private memory} (PRM) and \emph{shared memory} (SHM) used for device communication.
PRM is protected by the TEE and is not directly accessible to external devices.
In contrast, SHM lies outside the trusted memory boundary and serves as the explicit communication
region between the TEE and the external devices.
Therefore, data movement between the TEE and GPU follows the PRM-SHM-GPU data path. Specifically,
data is first copied from PRM to SHM through a TEE-side memory copy, and is then transferred from
SHM to GPU memory through Peripheral Component Interconnect Express (PCIe)~\cite{misono2024confidential}.
This path preserves the TEE protection boundary and defines the basic communication path for
TEE-GPU computation.

\subsection{Related Work}

{\bf TSDP for Transformer LLMs.}
Prior TSDP systems~\cite{tramerslalom,asvadishirehjini2020goat,shen2022soter,sun2023shadownet,zhang2024no}
mainly protect model and input confidentiality in partitioned TEE-GPU
execution.  
They encrypt or mask sensitive intermediate states before offloading
computation to an untrusted GPU, and then decrypt and verify the returned
results inside the TEE.
Recent systems extend this paradigm to
Transformer-based LLMs~\cite{wang2025game,xue2025securing}.
These TSDP systems do not consider the communication
overhead caused by moving protected states across the TEE-GPU boundary. 
In contrast,
our \ctga targets computation integrity and communication-efficient TEE-GPU
attention.

{\bf Side-channel attacks on TEEs.}
TEEs are known to be vulnerable to side-channel attacks. 
Such attacks primarily threaten confidentiality by leaking information 
from the protected execution domain, 
and may undermine integrity when they expose secrets used for
verification or authentication~\cite{yuan2024hypertheft,yuan2025ciphersteal}.
Various defenses have been proposed to mitigate side-channel leakage, 
including memory randomization schemes~\cite{li2022systematic,wichelmann2024obelix} 
and runtime profiling-based obfuscation~\cite{homescu2013profile}. 
\ctga can be combined with such defenses to reduce side channel leakage.

{\bf GPU TEE.}
Recent studies~\cite{volos2018graviton,wang2023building,sridhara2024acai} have explored trusted
architectures inside GPUs, including confidential-computing support on high-end GPUs such as
NVIDIA H100~\cite{nvidia_h100_2023}.
However, these approaches require specialized GPU hardware and firmware support. The related software
ecosystems are still maturing~\cite{wang2026confidential}.
As a result, they are not yet widely available on commodity GPUs and are less applicable to
edge clouds, where GPU resources are heterogeneous and cost-constrained~\cite{tian2025clone}.
In contrast, \ctga uses a VM-based CPU TEE as the root of trust and can work with any GPU.

%% file: Sections/Motivation.tex
\section{Motivation}
\label{sec:motivation}

\begin{figure}[!t]
	\centering
	\includegraphics[width=0.475\textwidth]{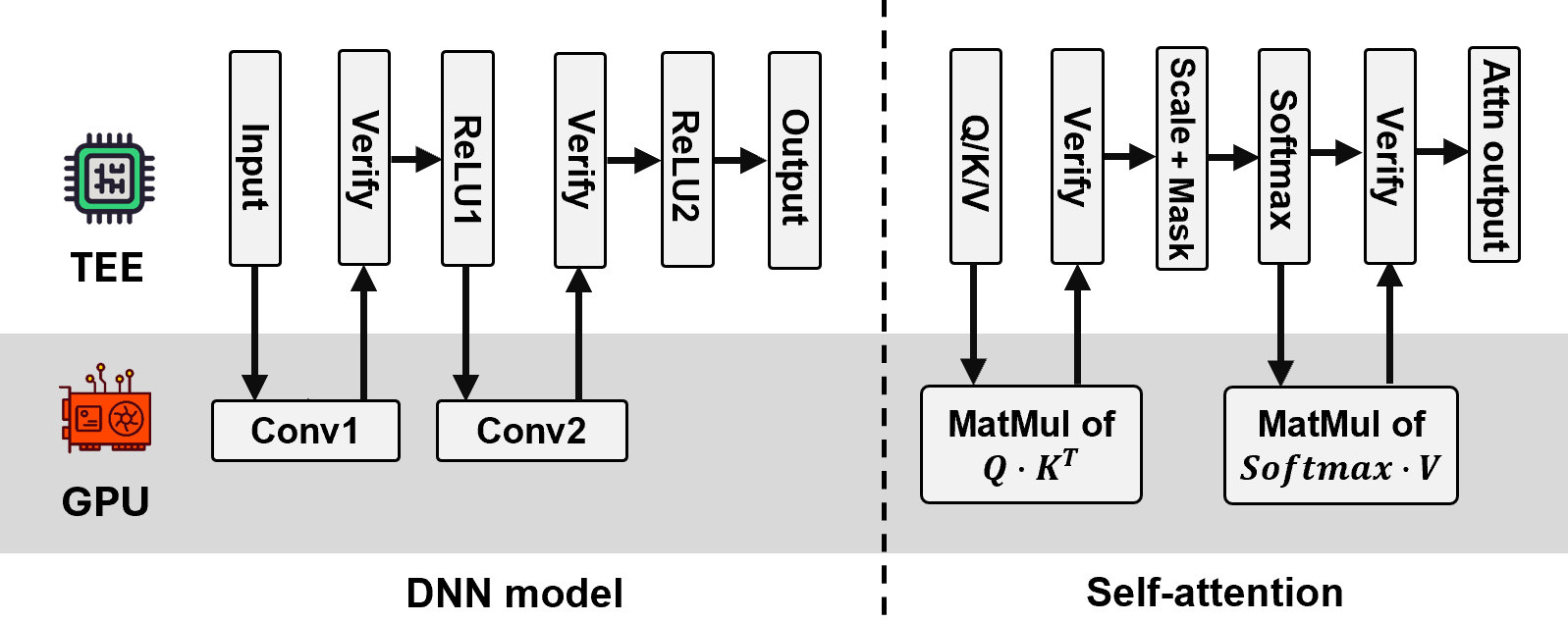}
    \vspace{-2em}
	\caption{Illustration of TSDP for a four-layer DNN model (left) and a Transformer self-attention module (right).}
	\label{fig:tsdp}
    \vspace{-1.2em}
\end{figure}

This section characterizes the bottlenecks of applying the sequential TSDP workflow
to Transformer self-attention in TEE-GPU inference.
TSDP offloads matrix multiplications to GPU,
while verification and non-linear operations remain in TEE. This design works well for conventional DNNs because their intermediate activations are relatively small
and their non-linear operators are lightweight.
The left part of Figure~\ref{fig:tsdp} illustrates implementing TSDP for a four-layer DNN model, where the convolutional layers are offloaded to the GPU. However, Transformer self-attention has different computation and communication patterns. The right part of Figure~\ref{fig:tsdp} illustrates a straw man proposal of applying TSDP to self-attention.
This section measures the performance of this straw man proposal with
two attention-level
microbenchmarks for the prefill and decoding phases, respectively, on the LLaMA3-8B~\cite{grattafiori2024llama} model with the
WikiText dataset~\cite{merity2016pointer}.

\begin{figure}[!t]
	\centering
	\includegraphics[width=0.47\textwidth]{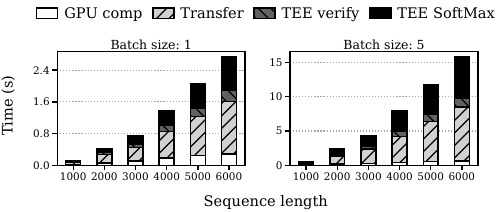}
    \vspace{-1em}
	\caption{Latency breakdown of TSDP prefill attention.}
	\label{fig:prefill_breakdown}
    \vspace{-1.2em}
\end{figure}

$\blacksquare$ {\bf Prefill bottleneck.} Figure~\ref{fig:prefill_breakdown} shows the latency breakdown of TSDP self-attention computation during prefill.
As sequence length grows from 1,000 to 6,000 tokens, 
attention state transfers and TEE-side \texttt{SoftMax} become major latency components.
Thus, the TEE-GPU communication overhead and TEE-side non-linear computation present the main bottlenecks to the straw man proposal.
This is because, self-attention produces large intermediate states,
whose sizes grow quadratically with the sequence length.
These intermediate states are repeatedly transferred across the TEE-GPU boundary, as illustrated in Figure~\ref{fig:tsdp}.
The serial TSDP workflow also suffers substantial idle times at the two processors, because the TEE and GPU alternately wait for
the states from each other.
Moreover, \texttt{SoftMax} is compute-intensive, as it requires
element-wise exponentiation and normalization over the full attention matrix.
As the sequence length increases, \texttt{SoftMax} overhead on TEE is more pronounced.

The results in Figure~\ref{fig:prefill_breakdown} suggest the necessity of offloading the main attention computation, including non-linear operations,
to GPU, in order to reduce TEE-GPU round trips and TEE-side computation overhead. They also motivate a pipelined workflow that overlaps TEE processing, data transfer, and GPU computation to reduce cross-device waiting and improve the utilization of both processors.

$\blacksquare$ {\bf Decoding bottleneck.}
Decoding has a different performance profile.
Each decoding step processes only one new query token, 
but it must attend to all previous tokens in the KV cache.
The KV cache grows with context length and concurrent requests, and can eventually exceed available GPU memory.
Modern LLM serving systems therefore offload inactive or overflowed KV-cache blocks to host memory~\cite{sheng2023flexgen,kwon2023efficient,lee2024infinigen,wang2025kvcache}.
In TEE-GPU attention, these non-resident KV entries reside in TEE-protected memory.
Decoding attention has low arithmetic intensity and is mainly limited by memory bandwidth on both
GPUs and TEE-enabled CPUs~\cite{he2024fastdecode,zhu2025nanoflow}.
Although CPUs are much weaker than GPUs for dense matrix computation, the performance
gap is smaller for decoding attention because the bottleneck is KV cache access rather than peak
compute throughput.
For example, an NVIDIA H20 GPU provides up to 4\,TB/s memory bandwidth and over 145\,TFLOPS of
compute throughput, whereas modern x86 server-class CPUs, including those used by VM-based TEEs, provide up to 500\,GB/s memory bandwidth specification in accessing multi-channel DDR5
memory, but only around 1.2\,TFLOPS of compute throughput~\cite{deng2025hgca,he2024fastdecode,zhu2025nanoflow}.
This narrower memory bandwidth gap suggests that computation inside the TEE can still be useful for
decoding attention.

\begin{figure}[!t]
	\centering
	\includegraphics[width=0.46\textwidth]{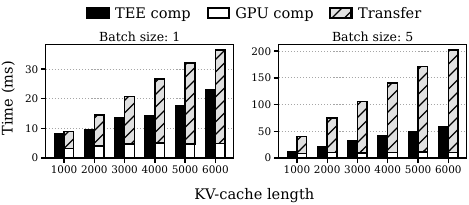}
    \vspace{-1em}
	\caption{Latency breakdown of TEE-based and GPU-based decoding attention.}
	\label{fig:kvtransfer}
    \vspace{-1.1em}
\end{figure}

Figure~\ref{fig:kvtransfer} compares TEE-based and GPU-based decoding attention 
when non-resident KV entries are stored in TEE-protected host memory.
The TEE-based strategy computes attention over these KV entries directly inside the TEE, 
whereas the GPU-based strategy first transfers the KV entries to GPU memory and then computes attention on the GPU.
Although GPUs provide higher compute throughput and memory bandwidth than CPUs, 
this advantage does not always translate to lower latency for single-token decoding.
In this setting, as decoding attention has low arithmetic intensity, GPU-based attention is dominated by the cost of transferring KV entries across the TEE-GPU boundary.
As a result, GPU-based attention can be slower than TEE-based attention 
despite the GPU's higher compute capability.

This observation motivates a collaborative TEE-GPU decoding workflow.
Instead of transferring all non-resident KV entries to the GPU, the TEE can compute partial attention
over KV entries already stored in protected memory, while the GPU processes resident and selected
offloaded KV blocks.

\begin{figure*}[!t]
	\centering
	\includegraphics[width=1\textwidth]{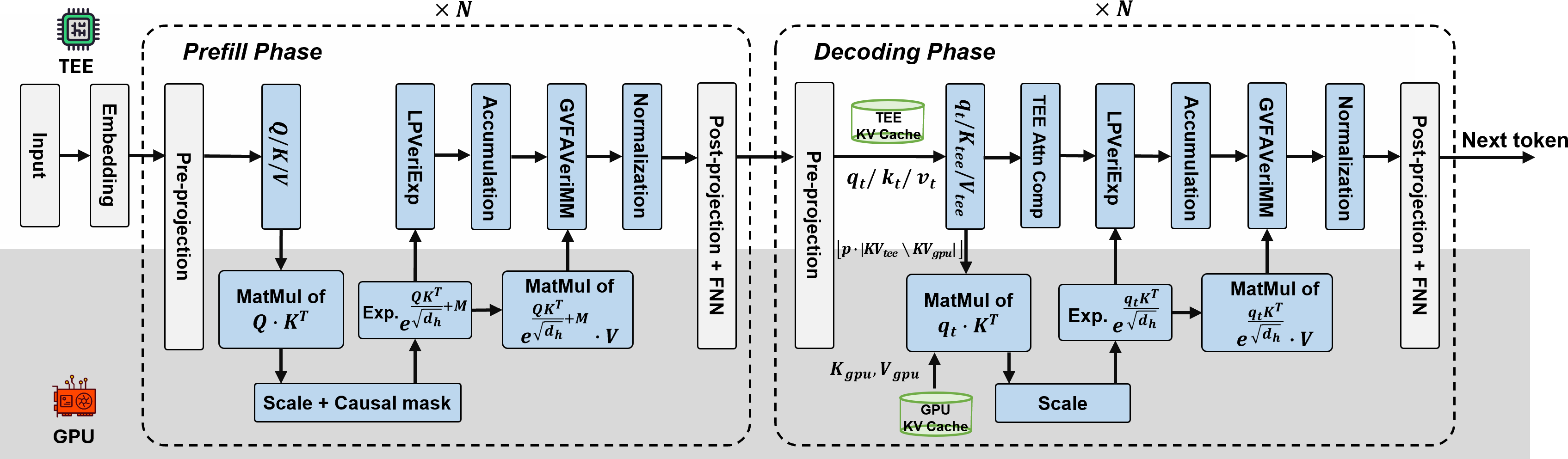}
    \vspace{-2em}
	\caption{Overview of \ctga for Transformer-based LLMs.}
	\label{fig:ctga}
    \vspace{-1.2em}
\end{figure*}

$\blacksquare$ {\bf TEE-GPU memory boundary.}
The above bottlenecks are further amplified by the memory boundary between the TEE and GPU.
As introduced in \sect\ref{subsec:tee_background}, TEE-GPU communication follows the
constrained PRM-SHM-GPU data path.
Data in PRM
cannot be directly used as a GPU direct memory access (DMA) buffer and must be staged through SHM.
Unlike ordinary CPU-GPU execution, where page-locked host memory enables efficient asynchronous DMA
transfers, TEE-GPU communication may involve page conversion, bounce buffers, or additional
host-side copies.
Recent work on GPU passthrough in Intel TDX reports that pageable transfers achieve only
$50$-$75\%$ of pinned-DMA throughput due to bounce-buffer overheads~\cite{sato2025towards}.
Therefore, even with unchanged physical PCIe bandwidth, the effective TEE-GPU path can suffer from
lower transfer efficiency, less overlap, and extra synchronization, further tightening the bottlenecks of
prefill state transfer and decoding KV cache transfer.
This highlights the need for a communication-efficient TEE-GPU attention framework
that reduces data movements and increases computation-communication overlap.

%% file: Sections/System_Design.tex
\section{\ctga System Design}
\label{sec:system_design}
This section presents \ctga,
designed to accelerate verifiable LLM inference 
by minimizing TEE-GPU communication
overhead and reducing attention computation latency.
Figure~\ref{fig:ctga} shows the overview of its architecture, 
which operates on the attention blocks of
LLM inference during both the prefill and decoding phases.

\subsection{Design Space}
\label{subsec:design_space}

{\bf Scope.}
\ctga targets verifiable LLM inference under the TEE-GPU setting, where the TEE
serves as the trusted component and the GPU is out of the trusted domain.
Model and input confidentiality are orthogonal to this work.
\ctga uses the GPU in an out-of-the-box manner and does not require specific GPU hardware,
firmware, or GPU TEE.
We focus on Transformer self-attention modules in both the prefill and decoding phases.
Other model components, such as projection and feed-forward layers, are dominated by linear operations 
and can adopt TSDP as shown in Figure~\ref{fig:tsdp}.

{\bf Threat model.}
Same as prior TEE-based verifiable offloading schemes
~\cite{tramerslalom,asvadishirehjini2020goat,asvadishirehjini2022ginn}, 
we consider two execution domains.
\begin{itemize}
    \item {\em TEE domain:}
    The TEE hardware and the attested \ctga code running inside the TEE are trusted.
    The TEE provides isolated memory for trusted code and protects the secret
    randomness used for verification.
    \item {\em GPU domain:}
    The GPU and its software stack are not trusted.
    They may deviate from the prescribed computation and return incorrect intermediate results.
\end{itemize}
All results returned from GPU are not trusted and must be verified inside the TEE
before being accepted.
We assume the adversary cannot break TEE isolation.
This paper does not consider denial-of-service and side-channel attacks.

{\bf Goals.} 
\ctga aims to make TEE-GPU attention computation both verifiable and efficient.
For integrity, it verifies GPU-returned results before they are used by trusted computation.
For efficiency, it reduces TEE-GPU communication volume and frequency, 
and idle time caused by sequential TEE-GPU execution.

\subsection{Attention Computation Verification}
\label{sec:attn_veri}
Figure~\ref{fig:ctga} shows how \ctga offloads both linear and non-linear
attention to GPU, including attention score computation, score
scaling, causal masking, exponentiation in \texttt{SoftMax}, and value
aggregation. With TSDP (cf. Figure~\ref{fig:tsdp}), GPU returns
intermediate attention scores to TEE for \texttt{SoftMax} computation and
then waits for the resulting attention weights to perform value
aggregation. In contrast, \ctga keeps both linear and non-linear attention
on GPU and returns only the intermediate results needed for
verification inside the TEE. 
Specifically, as illustrated in Figure~\ref{fig:ctga}, in both prefill and decoding, \ctga verifies the attention results returned by the GPU with procedures \texttt{GVFAVeriMM} and \texttt{LPVeriExp}.
We explain these two procedures in
\sect~\ref{subsubsec:linear-veri} and \sect~\ref{subsubsec:nonlinear-veri},
respectively, and provide their detailed pseudocode in
Appendix~\ref{app:Detailed-Verification-Procedures}.


\subsubsection{Linear computation verification}
\label{subsubsec:linear-veri}
We describe the approach to verifying offloaded linear computations.
TSDP systems
~\cite{tramerslalom,sun2023shadownet,zhang2024no,wang2025game}
typically verify matrix multiplication results returned by the GPU using
Freivalds' algorithm over a finite field~\cite{freivalds1977probabilistic}.
This requires TEE to quantize model weights and inputs into fixed-point
integer representations 
so that GPU execution preserves exact finite-field arithmetic.
However, in \ctga, the exponentiation offloaded to GPU is a real-valued non-linear operation that produces floating-point attention weights used by subsequent linear computations, rendering the Freivalds' algorithm inapplicable.
To address this, \ctga adopts the Gaussian
Variant of Freivalds' Algorithm (GVFA)~\cite{ji2020gaussian}, 
which is based on Lemma~\ref{lemma:GVFA}. The GVFA-based verification for matrix multiplication is 
referred to as \texttt{GVFAVeriMM}.

\begin{lemma}[Floating-point Gaussian Freivalds~\cite{ji2020gaussian}]
\label{lemma:GVFA}
Let $\mathbf{A}\in\mathbb{R}^{m\times n}$,
$\mathbf{B}\in\mathbb{R}^{n\times p}$, and
$\mathbf{C}\in\mathbb{R}^{m\times p}$ be arbitrary real matrices.
Let $\mathbf{w}_G\sim\mathcal{N}(0,I_p)$ be a standard Gaussian random vector,
for any tolerance $\epsilon>0$, 
\begin{equation}
\Pr\big[
\big\|(\mathbf{C}-\mathbf{A}\mathbf{B})\mathbf{w}_G\big\|_\infty
\le \epsilon
\mid
\mathbf{C}\neq\mathbf{A}\mathbf{B}
\big]
\le
2\Phi\big(\frac{\epsilon}{\sigma}\big)-1,
\end{equation}
where
$
    \sigma
    =
    \max_i
    \big\|
        (\mathbf{C}-\mathbf{A}\mathbf{B})_{i,:}
    \big\|_2,
$
and $\Phi(\cdot)$ is the cumulative distribution function of the standard
normal distribution.
\end{lemma}

Lemma~\ref{lemma:GVFA} provides a randomized consistency bound for
floating-point matrix multiplication.
The tolerance $\epsilon$ is calibrated from
honest GPU executions and specifies the accepted numerical error budget.
Results whose projected residuals fall within this budget are accepted
as numerically consistent with honest execution and treated as
normal floating-point variation.
This model is consistent with LLM quantization
studies showing that small controlled low-precision perturbations are often
tolerated by LLMs~\cite{frantar2022gptq,xiao2023smoothquant}.
We further evaluate the end-to-end impact of such tolerated numerical variation
in \sect~\ref{subsec:Verification-Effectiveness-and-Output-Quality}.
Deviations outside this budget are detected with the false-acceptance probability
bounded by Lemma~\ref{lemma:GVFA}.
In practice, the check would be repeated with $\tau$ independent Gaussian vectors,
which further reduces the false acceptance probability to
$\big(2\Phi(\epsilon/\sigma)-1\big)^\tau$.
\ctga uses this check for attention value aggregation in both prefill and
decoding. 
For prefill, following the notation in
\sect\ref{sec:background}, 
\ctga verifies whether the value aggregation
$\widehat{\mathbf U}$ returned by GPU is consistent with $\mathbf{E}\mathbf{V}$ within
tolerance $\epsilon$.
Instead of recomputing $\mathbf{E}\mathbf{V}$ inside the TEE, which requires
$O(L^2d_h)$ operations, for one repetition,
the TEE uses a secret Gaussian vector
$\mathbf{w}_G\in\mathbb{R}^{d_h}$ that is sampled and stored during the
offline phase
\begin{equation}
    \big\|
        \mathbf{E}(\mathbf{V}\mathbf{w}_G)
        -
        \widehat{\mathbf U}\mathbf{w}_G
    \big\|_\infty
    \le
    \epsilon.
    \label{eq:linearcheck-prefill}
\end{equation}
This reduces verification to matrix-vector products with an overall complexity of
$O(L^2+Ld_h)$.

For decoding, each step verifies a vector-matrix product.
At step $t$, 
given the GPU-returned claim $\widehat{\mathbf u}_t$,
the TEE checks
\begin{equation}
    \big|
        \mathbf{e}_t(\mathbf{V}_{\le t}\mathbf{w}_G)
        -
        \widehat{\mathbf u}_t\mathbf{w}_G
    \big|
    \le
    \epsilon.
    \label{eq:linearcheck-decoding}
\end{equation}
A direct computation of
$\mathbf{V}_{\le t}\mathbf{w}_G$ costs $O(td_h)$ per step.
Since decoding is autoregressive, \ctga maintains this projection
incrementally.
For a fixed Gaussian vector $\mathbf{w}_G$, the TEE stores
$\mathbf{g}_{t-1}=\mathbf{V}_{\le t-1}\mathbf{w}_G$.
When a new value vector $\mathbf{v}_t$ is appended to the KV cache, the TEE
updates
$    \mathbf{g}_t
    =
    \mathbf{V}_{\le t}\mathbf{w}_G
    =
    \begin{bmatrix}
        \mathbf{g}_{t-1}\\
        \mathbf{v}_t^{\top}\mathbf{w}_G
    \end{bmatrix}$.
Thus, decoding verification requires $O(t+d_h)$ operations for the check in~\eqref{eq:linearcheck-decoding}, 
compared with $O(td_h)$ for recomputing $\mathbf{u}_t$.

\subsubsection{Non-linear computation verification}
\label{subsubsec:nonlinear-veri}
This section describes how to verify exponentiation in \texttt{SoftMax}.
Freivalds' algorithm and GVFA rely on algebraic properties of linear
operations and cannot be applied to element-wise non-linear
operators.
To verify exponentiation, we propose \texttt{LPVeriExp}, which uses a
log-product consistency property of the exponential function. Specifically,
consider the $r$-th row of the attention-score matrix during prefill. 
Let $s_{r,i}$ be the $(r,i)$-th element of attention
score $\mathbf{S}$, where
$i\in\mathcal I_r$ and $\mathcal I_r$ denotes the valid, unmasked positions in
the $r$-th row. 
The exact real-valued exponentiation output is $y_{r,i}=e^{s_{r,i}}$.
The GPU returns a claimed exponentiation vector
$
    \widehat{\mathbf y}_r=[\widehat y_{r,i}]_{i\in\mathcal I_r}.
$
For each verification repetition, the TEE uses a global coefficient vector
$
    \mathbf a=[a_i]_{1\leq i \leq L}
$,
$
    a_i \in \{1,\ldots,N_a\},
$
and an independent Gaussian scalar $w'_{G} \sim\mathcal N(0,1)$,
where $N_a$ is the size of the nonzero coefficient domain. 
Both $\mathbf a$ and $w'_G$ are sampled during the offline phase 
inside the TEE and kept hidden from the GPU.
For exact exponentiation outputs, we have
$    \log\big(
        \prod_{i\in\mathcal I_r}
        (y_{r,i})^{a_i}
    \big)
    =
    \log\big(
        e^{\sum_{i\in\mathcal I_r}a_i s_{r,i}}
    \big)                                                       
    =
    \sum_{i\in\mathcal I_r}a_i s_{r,i}$.
Next, \texttt{LPVeriExp} computes the scalar residual
\begin{equation}
    R_r(\mathbf a)
    =
    \log\big(
        \prod_{i\in\mathcal I_r}
        \widehat y_{r,i}^{a_i}
    \big)
    -
    \sum_{i\in\mathcal I_r}a_i s_{r,i}.
    \label{eq:LPVerifyExp-residual}
\end{equation}
Because floating-point arithmetic can introduce small numerical residuals
even for honest GPU execution, the TEE uses a calibrated tolerance
$\epsilon'$.
Then, \texttt{LPVeriExp}
checks whether 
\begin{equation}
    |R_r(\mathbf a)w'_{G}|<\epsilon'.
    \label{eq:lpcheck}
\end{equation}
This log-product verification binds the returned exponentiation outputs to the
preceding computation.
Since each $s_{r,i}$ is derived from $\mathbf{Q}\mathbf{K}^\top$, score scaling, and causal
masking, an incorrect score or exponentiation output introduces an additional
log-product residual.
The following lemma bounds the false acceptance probability of the check in~\eqref{eq:lpcheck}.

\begin{lemma}[False Acceptance of \texttt{LPVeriExp}]
\label{lemma:lpverifyexp}
For the $r$-th attention row, let $n_{\mathrm{err},r}$ denote the number of exponentiation outputs 
modified from an honest GPU execution
and 
$\sigma'>0$ denote the minimum detectable magnitude 
of an incorrect verification residual.
With $\tau_a$ fresh secret integer vectors
and $\tau_g$ independent Gaussian tolerance checks
per coefficient vector,
the probability that an incorrect row is accepted is bounded by
\begin{equation}
\Pr[\mathsf{FA}_r] \le 
\big( \mathbf 1\{n_{\mathrm{err},r}\ge2\} 
\cdot \frac{1}{N_a} + 
\big( 2\Phi\big(\frac{\epsilon'}{\sigma'}\big)-1 
\big)^{\tau_g} 
\big)^{\tau_a}.
\end{equation}
\end{lemma}
By Lemma~\ref{lemma:lpverifyexp}, increasing $N_a$, $\tau_a$, and $\tau_g$
reduces the false acceptance probability and can make it sufficiently small.
We provide the proof in Appendix~\ref{app:proof-nonlinear-verification},
where we also describe a numerically stable implementation of the
log-product check in~\eqref{eq:LPVerifyExp-residual} that avoids intermediate
underflow and overflow.

We now compare the verification cost with directly executing the score
computation and exponentiation inside the TEE.
Direct TEE-side execution requires $O(L^2d_h)$ linear operations for
$\mathbf Q\mathbf K^\top$ and $O(L^2)$ expensive exponentiations.
In contrast, for computing the residual in~\eqref{eq:LPVerifyExp-residual},
\texttt{LPVeriExp}   first computes the compressed score term for each
row by
$    \sum_{i\in\mathcal I_r}a_i s_{r,i}
    =
        \mathbf q_r^\top
        \big(
            \sum_{i\in\mathcal I_r}a_i\mathbf k_i
        \big)
/\sqrt{d_h}
    +
    \sum_{i\in\mathcal I_r}a_i\mathbf M_{r,i}$,
where $\mathbf q_r^\top$ and $\mathbf k_i^\top$ are the corresponding query and
key rows.
For a fixed vector $\mathbf a$, 
the weighted key terms
$
    \sum_{i\in\mathcal I_r}a_i\mathbf k_i
$
can be computed incrementally and reused across rows.
Thus, computing all compressed score terms takes $O(Ld_h+L^2)$ linear
operations.
For the log-product term in~\eqref{eq:LPVerifyExp-residual}, the TEE computes
integer powers $\widehat y_{r,i}^{a_i}$ using exponentiation by squaring.
This takes $O(\log N_a)$ multiplications per element and thus
$O(L^2\log N_a)$ linear operations over all exponentiation outputs.
The only remaining non-linear operations are one logarithm per row, for a total of $O(L)$ logarithmic operations.

During prefill, $L>d_h$ typically holds
as \(d_h\) is usually $64$ or $128$~\cite{grattafiori2024llama,yang2025qwen3},
and $\log N_a$ is much smaller than $d_h$.
Therefore, compared with direct TEE execution, \texttt{LPVeriExp} 
reduces the linear computation from $O(L^2d_h)$ to
$
    O(Ld_h+L^2+L^2\log N_a)
$
and reduces expensive non-linear operations from $O(L^2)$ exponentiations to
$O(L)$ logarithms. Note that the computational overheads of logarithm and exponentiation are comparable.
Similarly, during decoding, the same strategy is applied only 
to the current token attention row. At step $t$, direct TEE execution would
require $O(td_h)$ linear operations and $O(t)$ exponentiations, whereas
\texttt{LPVeriExp} requires $O(td_h+t\log N_a)$ linear operations   
and only one logarithm, i.e., $O(1)$ nonlinear operation.
Therefore, \texttt{LPVeriExp} is substantially more efficient in both phases 
than directly performing the computation inside the TEE.

\subsection{Prefill Workflow}
\label{subsec:prefill_stage}
We now describe the prefill workflow of \ctga, which integrates the verification mechanisms introduced in \sect\ref{sec:attn_veri}. 
As discussed earlier, TSDP keeps \texttt{SoftMax} inside the TEE and requires quadratic-size attention states to cross the TEE-GPU boundary before and after \texttt{SoftMax}. 
With \texttt{LPVeriExp} and \texttt{GVFAVeriMM}, 
\ctga removes this related round trip.
However, existing TSDP offloading schemes
still follow a sequential workflow: 
TEE and GPU cannot proceed until the other side finishes its data transfer
or computation, which can leave both sides idle. 
This issue can be pronounced because the exponentiation outputs are still quadratic in the
sequence length and must be returned to the TEE for verification.

\begin{figure}[!t]
	\centering
	\includegraphics[width=0.478\textwidth]{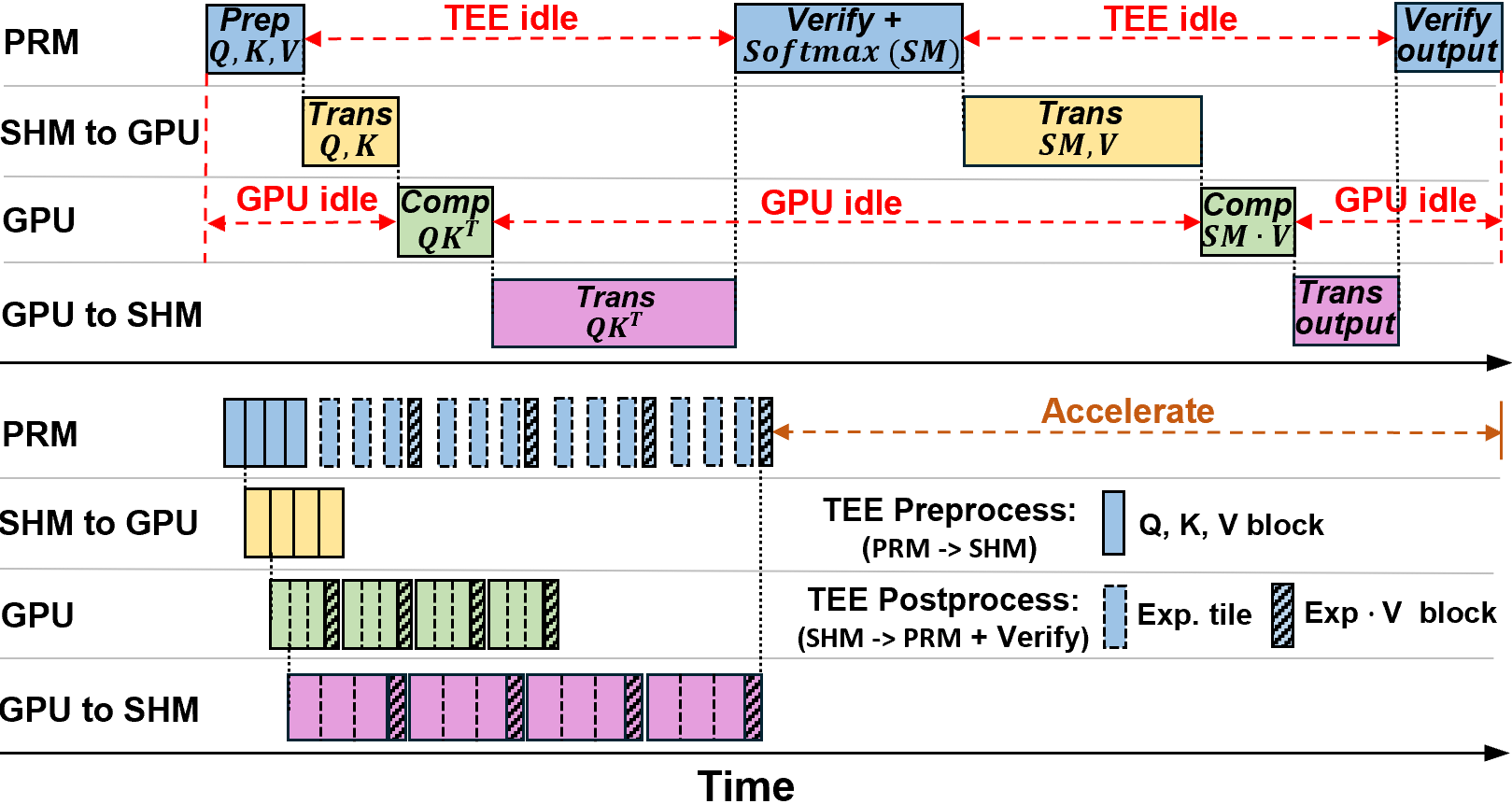}
    \vspace{-2em}
	\caption{Comparison of the TSDP sequential workflow (top) and \ctga pipelined workflow (bottom) in prefill.}
	\label{fig:serial_workflow_prefill}
    \vspace{-1.2em}
\end{figure}

Inspired by distributed DNN/LLM pipelines that overlap communication with
computation and reduce pipeline bubbles across GPU stages~\cite{narayanan2019pipedream,narayanan2021efficient,sun2025seq1f1b},
\ctga applies pipelining inside attention computation at the TEE-GPU boundary.
It uses a two-level pipeline over attention head blocks and exponentiation row tiles to overlap TEE preprocessing and postprocessing, SHM-GPU communication, and GPU computation. 
Preprocessing copies input blocks from PRM to SHM.
Postprocessing copies GPU results from SHM to PRM, verifies them using \texttt{LPVeriExp} and \texttt{GVFAVeriMM}, accumulates row sums, and normalizes the verified results.
Figure~\ref{fig:serial_workflow_prefill} compares the TSDP sequential workflow with the \ctga pipelined workflow. We describe the details below.

$\blacksquare$ {\bf Head block pipeline.}
At a coarse granularity, \ctga exploits the independence across attention
heads and groups them into head blocks. 
Let $\mathcal{B}\gets\{1,\ldots,\lceil H/b_s\rceil\}$ denote the set of head blocks, where $b_s$ is the number of heads in each block. 
Each block $b\in\mathcal{B}$ contains up to $b_s$ heads and the corresponding tensors $\mathbf{Q}_{b}$, $\mathbf{K}_{b}$, and $\mathbf{V}_{b}$.
While the TEE copies the next head block from PRM to SHM, the GPU transfers a
previously ready block from SHM to GPU memory. 
Once the block arrives on the
GPU, the GPU starts to compute the attention scores
$\mathbf{S}_b=\mathbf{Q}_b\mathbf{K}_b^\top/\sqrt{d_h}+\mathbf{M}_b$,
while later blocks are being transferred. 
This overlaps PRM-SHM memory copy, SHM-GPU PCIe transfer, and
GPU computation across head blocks, improving the utilization of
both the TEE and GPU.

$\blacksquare$ {\bf Row tile pipeline.}
At a finer granularity, \ctga further partitions the quadratic-size
exponentiation output of each head block $b$ into
row tiles. 
Let $\mathcal{T}\gets\{1,\ldots,\lceil L/t_s\rceil\}$ denote the set of
row tiles where $t_s$ is the number of rows in each tile.
Since exponentiation is element-wise, 
each tile $s \in \mathcal{T}$ can be processed
independently. 
Once the GPU finishes computing a tile, 
it transfers the claimed exponentiation output tile $\widehat{\mathbf{E}}_{b,s}$ to SHM. 
The TEE next copies the tile to PRM after it arrives in SHM,
verifies it using \texttt{LPVeriExp},
and accepts it as
$\mathbf{E}_{b,s}$ if the check passes. 
The TEE then accumulates the corresponding row-sum tile
$\mathbf{Z}_{b,s}=\mathbf{E}_{b,s}\mathbf{1}.$
These stages
are pipelined across different tiles: while the TEE postprocesses an earlier
tile, PCIe transfer can move another ready tile to SHM, and the GPU can compute
later tiles. This enables an overlap among TEE postprocessing, exponentiation transfer, and GPU computation, 
allowing verification to
start before the full exponentiation matrix of the head block is available.

After verifying all exponentiation tiles in head block $b$, 
the TEE concatenates them into the verified exponentiation matrix $\mathbf{E}_b$ and its row-sum vector $\mathbf{Z}_b=\mathbf{E}_b\mathbf{1}$.
Unlike exponentiation, value aggregation is verified at the granularity of an entire head block, because \texttt{GVFAVeriMM} performs a matrix-level check that requires the complete input and output matrices. After the GPU finishes value aggregation for block $b$, it returns a claimed result $\widehat{\mathbf{U}}_b$. The TEE verifies $\widehat{\mathbf{U}}_b$ using \texttt{GVFAVeriMM} with the accepted $\mathbf{E}_b$ and $\mathbf{V}_b$. If the check passes, the result is accepted as $\mathbf{U}_b$, and the TEE computes the normalized output $\mathbf{O}_b=\mathbf{U}_b\oslash\mathbf{Z}_b$.
Concatenating all
head-block outputs gives the final attention output.
Pseudocode for
the prefill workflow is provided in Appendix~\ref{app:prefill-workflow}.

\subsection{Decoding Workflow}
\label{subsec:decoding_stage}
\begin{figure}[!t]
	\centering
	\includegraphics[width=0.478\textwidth]{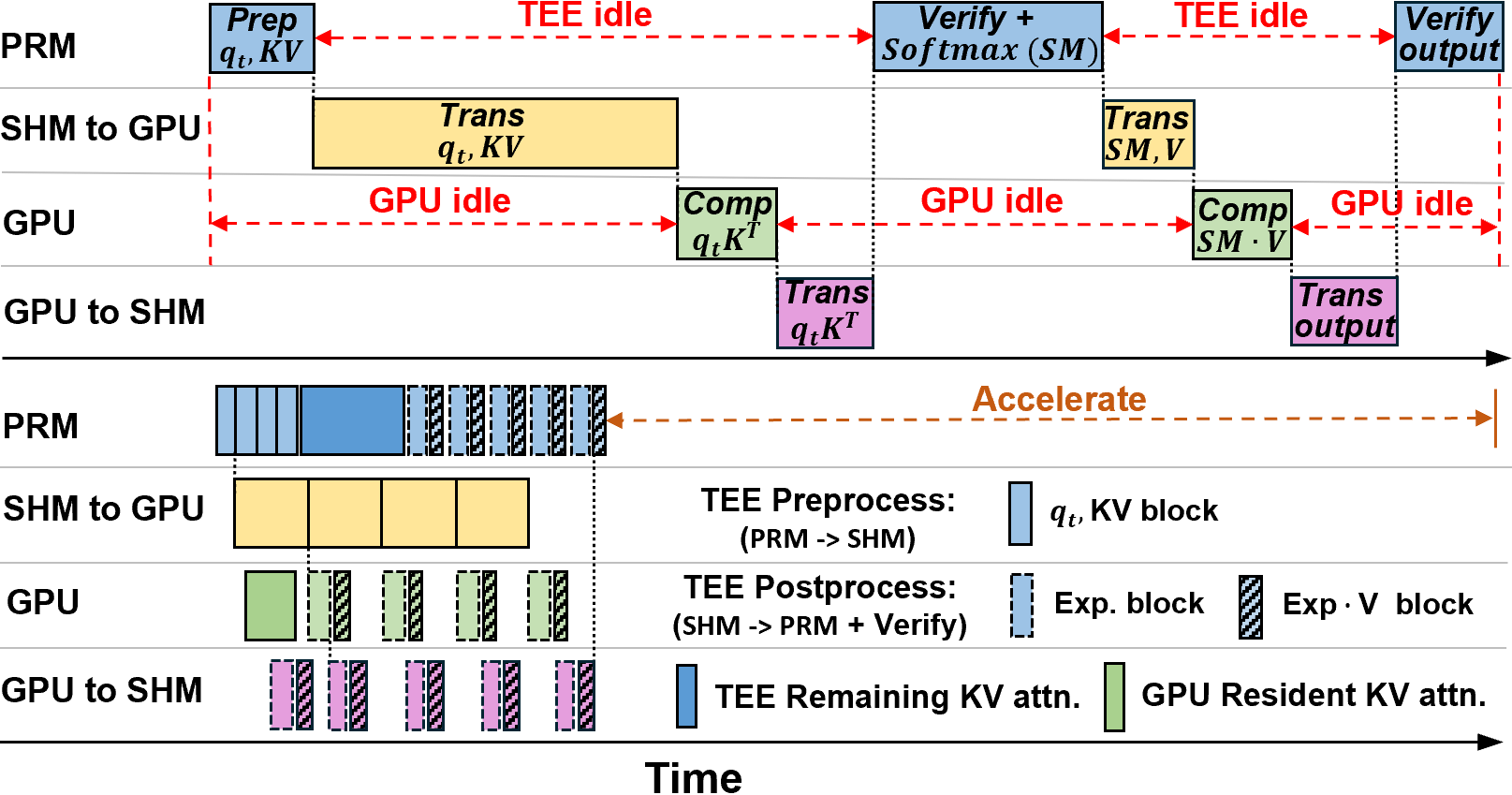}
    \vspace{-2em}
	\caption{Comparison of TSDP sequential workflow (top) and \ctga collaborative workflow (bottom)
    in decoding.}
	\label{fig:serial_workflow_decoding}
    \vspace{-1.2em}
\end{figure}

This section describes the decoding workflow of \ctga.
Unlike prefill,
decoding processes one query token at a time, but each step must attend to the
growing KV cache. When the KV cache exceeds GPU memory capacity, some KV entries
only remain in TEE memory. A decoding path that computes attention only on the
GPU must repeatedly move these entries through the PRM-SHM-GPU path. 
This transfer can dominate latency because each decoding
step accesses a large amount of KV data but performs limited computation, making
the computation insufficient to amortize KV
transfer overhead.
To address this,
\ctga 
introduces a collaborative decoding workflow that leverages the computational capability of the TEE and effectively hides transfer latency.
Figure~\ref{fig:serial_workflow_decoding} compares the TSDP sequential workflow with the collaborative workflow in \ctga.

$\blacksquare$ {\bf Collaborative attention.}
In contrast to GPU-centric LLM serving and offloading systems that focus on improving GPU utilization and memory efficiency~\cite{yu2022orca,kwon2023efficient,sheng2023flexgen,agrawal2023sarathi,zhong2024distserve},
\ctga splits the attention computation across the TEE and GPU for long-context decoding.
The TEE keeps the complete KV cache $\mathbf{KV}_{tee}$,
while the GPU maintains bounded resident KV cache $\mathbf{KV}_{gpu}$
for fast access. 
At step $t$, the model generates $\mathbf{q}_t$, $\mathbf{k}_t$, and $\mathbf{v}_t$, 
and appends $\mathbf{k}_t$ and $\mathbf{v}_t$ to the TEE-resident KV cache.
If the GPU still has available memory, the new KV entry is also copied and appended to $\mathbf{KV}_{gpu}$ 
for use in subsequent decoding steps.
When $\mathbf{KV}_{gpu}$ reaches its capacity, 
later KV entries are kept only in the TEE-resident KV cache unless they are temporarily selected for GPU execution.

\ctga uses an offload ratio $p \in[0,1]$ to partition 
the KV entries $\mathbf{KV}_{tee} \backslash \mathbf{KV}_{gpu}$ 
that are not resident on the GPU.
It streams 
$\lfloor p\cdot|\mathbf{KV}_{tee}\backslash\mathbf{KV}_{gpu}|\rfloor$ entries
to the GPU and processes the remaining KV entries 
inside the TEE.
Increasing $p$ improves GPU utilization but raises TEE-GPU transfer, 
while decreasing $p$ reduces KV
movement but assigns more computation to the TEE. 
Thus, $p$ tunes the trade-off
among GPU acceleration, TEE-side computation, and transfer overhead.

During decoding, the GPU computes attention over its resident KV cache and selected non-resident KV entries, 
while the TEE computes attention over the remaining KV entries in PRM.
To avoid exceeding the GPU KV capacity, \ctga partitions the selected KV entries into blocks and streams them to the GPU 
through a reusable temporary buffer.
The two paths overlap data movement and computation: 
the TEE copies the current query and selected KV blocks from PRM to SHM and performs local attention, 
while the GPU transfers each block from SHM to the temporary buffer, computes linear and non-linear partial attention, and returns intermediate results to SHM for TEE-side verification. 
This design reduces repeated KV transfers and allows TEE and GPU workloads to run concurrently 
while keeping GPU memory usage bounded.

$\blacksquare$ {\bf Partial state merge.} 
Both sides compute attention in an unnormalized form. Let $(\mathbf{U}_{gpu},\mathbf{Z}_{gpu})$ and $(\mathbf{U}_{tee},\mathbf{Z}_{tee})$ denote the partial states computed over the KV partitions on the GPU and TEE, respectively. Similar to the prefill workflow, the TEE copies the claimed GPU results for exponentiation and value aggregation from SHM to PRM, and verifies them using \texttt{LPVeriExp} and \texttt{GVFAVeriMM}.
The TEE then merges the two partial states to obtain the attention output: 
$
    \mathbf{o}_t =
    \frac{\mathbf{U}_{gpu}+\mathbf{U}_{tee}}
        {\mathbf{Z}_{gpu}+\mathbf{Z}_{tee}}.
$
This is equivalent to full attention over the entire KV cache because both the unnormalized numerator and the normalization denominator are additive across disjoint KV partitions.
If partition-local max-shifted softmax is used for numerical stability, \ctga
applies the standard log-sum-exp correction before merging.
Detailed pseudocode for the decoding workflow is provided in Appendix~\ref{app:decoding_workflow}.

%% file: Sections/Evaluation.tex
\section{Evaluation}
\label{sec:eval}

\begin{figure}[!b]
\vspace{-0.7em}
	\centering
	\includegraphics[width=0.42\textwidth]{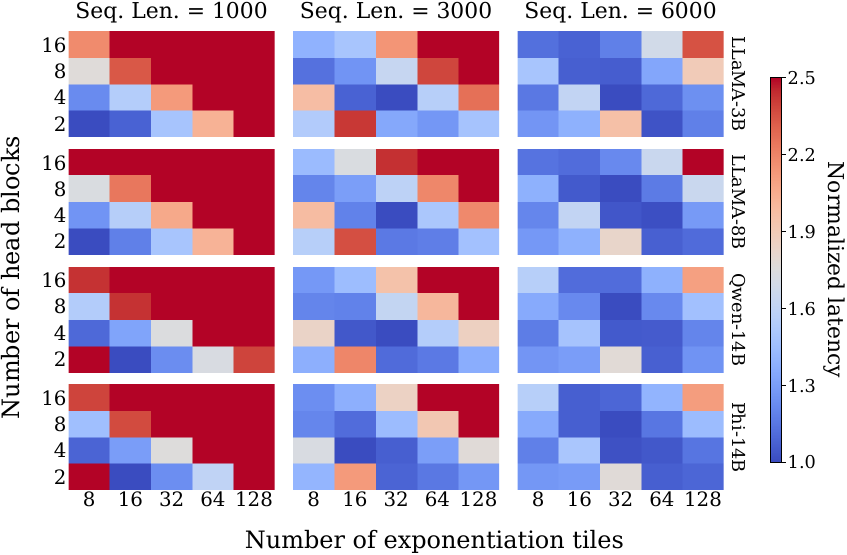}
    \vspace{-1em}
    \caption{Normalized prefill attention latency under different block and tile counts.}
    \label{fig:prefill-calibration}
\end{figure}

\begin{figure*}
	\centering
	\includegraphics[width=1\textwidth]{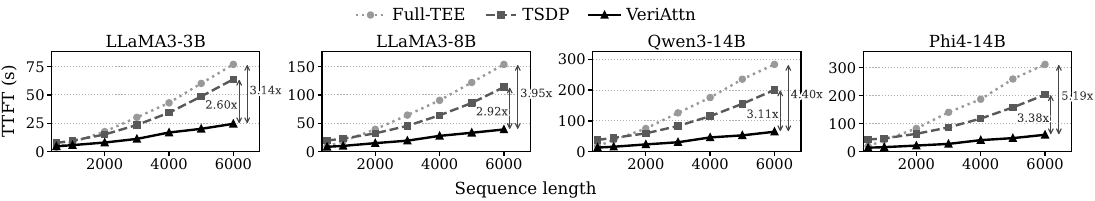}
    \vspace{-2em}
	\caption{Prefill TTFT comparison among Full-TEE, TSDP, and \ctga across four models under varying sequence lengths.}
	\label{fig:prefill_ttft}
    \vspace{-0.8em}
\end{figure*}

\subsection{Experimental Setup}
\label{subsec:eval_setup}

{\bf Testbed.}
All experiments were conducted in a TDX-enabled cloud VM hosted on an Intel Xeon Scalable CPU.
The VM has $16$ vCPUs, 128\,GB memory, and access to an NVIDIA H20 GPU with 96\,GB device memory through PCIe~5.0.
The effective single-thread PRM-SHM copy bandwidth is up to $32$~GB/s, and the SHM-GPU
bandwidth is about $15$~GB/s.
The end-to-end pageable transfer bandwidth remains below the PCIe~5.0 peak because PRM data must be staged through SHM, whose mapping cannot be registered as CUDA-pinned memory inside the TDX VM.

{\bf Models and dataset.}
We use four open-weight and representative large Transformer models 
with varying sizes and architectural designs: 
LLaMA3-3B~\cite{meta2024llama32},
LLaMA3-8B~\cite{grattafiori2024llama},
Qwen3-14B~\cite{yang2025qwen3}, 
and Phi4-14B~\cite{abdin2024phi}.
Unless otherwise stated, models by default are loaded in FP16 precision.
To evaluate inference performance, 
we use the WikiText dataset~\cite{merity2016pointer}, 
which provides natural text sequences for constructing prefill prompts 
and evaluating long-context decoding workloads.

{\bf Baselines.}
We compare \ctga against three representative approaches for verifiable LLM inference:
\textup{(1)} \textbf{Full-TEE}, running the entire model inside the TEE without GPU offloading;
\textup{(2)} \textbf{TSDP}, following the sequential TEE-shielded DNN partition workflow adopted by prior systems~\cite{tramerslalom,sun2023shadownet,zhang2024no,wang2025game};
and \textup{(3)} \textbf{zkLLM}~\cite{sun2024zkllm}, verifying offloaded LLM computation using cryptographic proofs.

\begin{figure}[!t]
\centering
\begin{minipage}[t]{0.235\textwidth}
\vspace{0pt}
\centering
\includegraphics[
width=\linewidth,
height=2.2cm
]{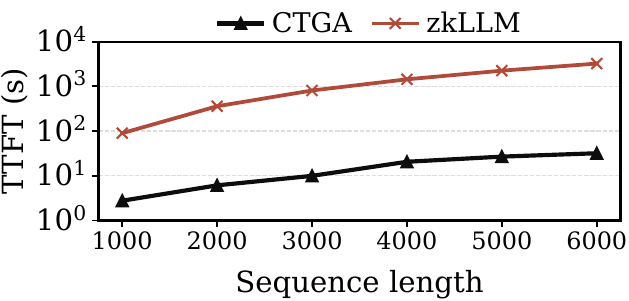}
\vspace{-2em}
\caption{\ctga vs. zkLLM in prefill TTFT on LLaMA3-8B.}
\label{fig:llama8b_prefill_attention}
\end{minipage}
\hfill
\begin{minipage}[t]{0.235\textwidth}
\vspace{0pt}
\centering
\includegraphics[
width=\linewidth,
height=2.2cm
]{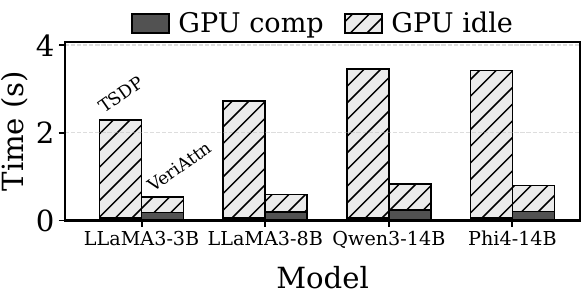}
\vspace{-2em}
\caption{GPU latency breakdown (prefill attention).}
\label{fig:prefill-gpu-breakdown}
\end{minipage}
\vspace{-1em}
\end{figure}

\subsection{Prefill Performance}
\label{sec:eval-prefill-calibration}
\subsubsection*{Sensitivity to block and tile counts.}
We first study the impact of two pipeline scheduling parameters 
on prefill self-attention latency of \ctga:
the number of attention head blocks and the number of exponentiation tiles.
These two parameters determine the granularity of \ctga's two-level prefill pipeline:
head blocks define the outer head-level pipeline; exponentiation tiles 
define the inner tile-level pipeline for transferring and verifying exponentiation outputs.
For each model, 
we sweep the number of head blocks over $\{2,4,8,16\}$ and the number of
exponentiation tiles over $\{8,16,32,64,128\}$ under sequence lengths of
1,000, 3,000, and 6,000 tokens.
For each model and sequence length, we normalize every measured latency by the minimum latency in
the same sweep.
Thus, a normalized latency of $1.0$ denotes the fastest configuration, 
and larger values indicate higher relative overhead.
Figure~\ref{fig:prefill-calibration} shows a clear trade-off that depends on sequence length.
Short prompts favor coarse settings, because excessive head partitioning or exponentiation tiling
introduces noticeable scheduling overhead.
For longer prompts, moderate partitioning improves pipeline parallelism while keeping enough work in
each head block and exponentiation tile for efficient GPU execution.
Guided by these results, 
we use length-aware defaults in subsequent prefill experiments:
$(2,16)$ head blocks and exponentiation tiles for sequence lengths up to 1,000,
$(4,32)$ for sequence lengths up to 3,000, and $(8,32)$ for longer sequences.

\subsubsection*{TTFT in the prefill phase.}
Figure~\ref{fig:prefill_ttft} compares prefill Time-To-First-Token (TTFT) for a single request, i.e., batch size $1$, 
across Full-TEE, TSDP, and \ctga.
We evaluate four models with sequence lengths ranging from 1,000 to 6,000 tokens.
\ctga{} consistently outperforms both baselines across all evaluated sequence lengths.
At a sequence length of 6,000, \ctga{} achieves speedups of
$2.60$-$3.38\times$ over TSDP and $3.14$-$5.19\times$ over Full-TEE.
These results demonstrate that \ctga effectively reduces prefill latency with verification enabled.
We also compare \ctga with zkLLM-style proof generation on LLaMA3-8B, as shown in
Figure~\ref{fig:llama8b_prefill_attention}.
The results show that zkLLM incurs roughly two orders of magnitude higher latency than \ctga.
This is because \ctga avoids expensive proof generation and instead uses lightweight TEE-side
verification for GPU-returned results.

\begin{figure}[!t]
	\centering
	\includegraphics[width=0.46\textwidth,height=6.8cm]{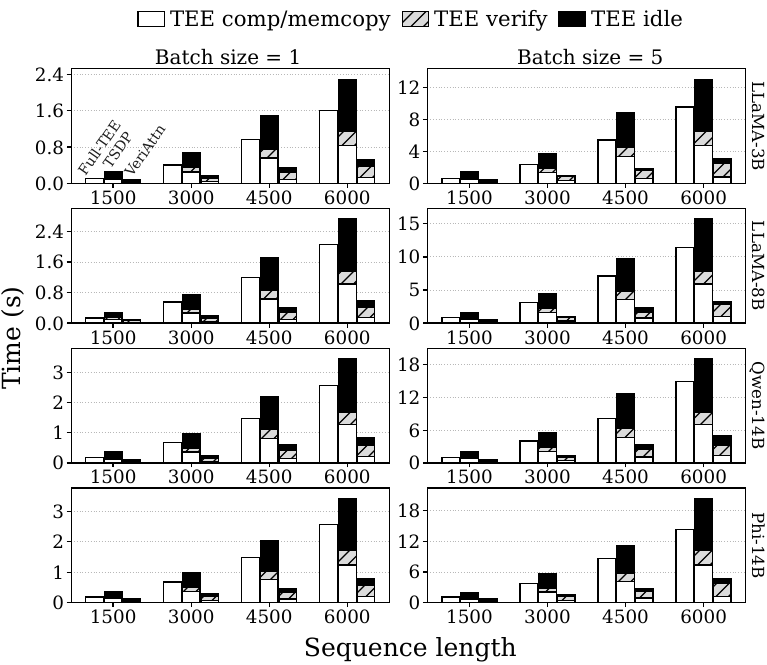}
    \vspace{-1em}
    \caption{
     TEE latency breakdown of prefill attention.
    }
    \label{fig:prefill-latency-Breakdown-ablation}
    \vspace{-1em}
\end{figure}

\subsubsection*{Prefill attention latency breakdown.}
Figure~\ref{fig:prefill-latency-Breakdown-ablation} reports the TEE-side latency breakdown of
prefill self-attention for Full-TEE, TSDP, and \ctga across four models, two batch sizes, and
sequence lengths from 1,500 to 6,000 tokens.
This experiment isolates the self-attention operator and excludes non-attention layers such as
feed-forward layers, layer normalization, and projections.
Each bar is decomposed into three components:
\textit{TEE comp/memcopy} includes TEE-side attention computation and PRM-SHM memory-copy overhead;
\textit{TEE verify} is the cost of verifying GPU-returned intermediate results; and
\textit{TEE idle} is the time the TEE waits for GPU computation and SHM-GPU data movement.

Full-TEE is dominated by TEE-side computation, as the entire attention operation runs inside the TEE.
TSDP offloads linear attention operations to the GPU, but keeps \texttt{SoftMax} inside the TEE and
uses a sequential TEE-GPU workflow.
Unlike the full-model TTFT results in Figure~\ref{fig:prefill_ttft}, 
TSDP can be slower than Full-TEE when attention is isolated.
This is because TSDP is effective for non-attention linear layers, where the TEE handles only
lightweight non-linear work and transfers relatively small intermediate states.
In self-attention, however, \texttt{SoftMax} is costly inside the TEE, and the intermediate
attention states are much larger.
Sequential TEE-GPU transfers further increase both communication delay and TEE idle time, while
TEE-side \texttt{SoftMax} adds substantial computation overhead.
In contrast, \ctga offloads both non-linear exponentiation and linear value aggregation to the GPU
and returns only the intermediate results needed for verification inside the TEE.
This reduces both TEE-side computation and TEE-GPU transfer overhead.
Its two-level pipeline overlaps GPU computation, data movement, and TEE-side verification,
which further reduces TEE idle time.
The benefit grows with sequence length.
For isolated prefill self-attention at 6,000 tokens and batch size 1, 
\ctga is $3.05$-$3.49\times$ faster than Full-TEE 
and $4.19$-$4.62\times$ faster than TSDP 
across the four models.
At batch size 5, \ctga shows a similar advantage, achieving
$3.02$-$3.52\times$ speedup over Full-TEE and
$3.87$-$4.84\times$ speedup over TSDP.
Figure~\ref{fig:prefill-gpu-breakdown} 
gives the complementary GPU-side view at sequence length
6,000 and batch size 1.
Compared with TSDP, \ctga spends a larger portion of time in useful GPU computation because more
attention work is offloaded to the GPU.
Also, \ctga substantially reduces GPU idle time across all four models, showing that the
pipeline keeps the GPU better utilized instead of leaving it stalled by sequential transfers and
TEE-side \texttt{SoftMax}.

\begin{figure}[!t]
	\centering
	\includegraphics[width=0.44\textwidth]{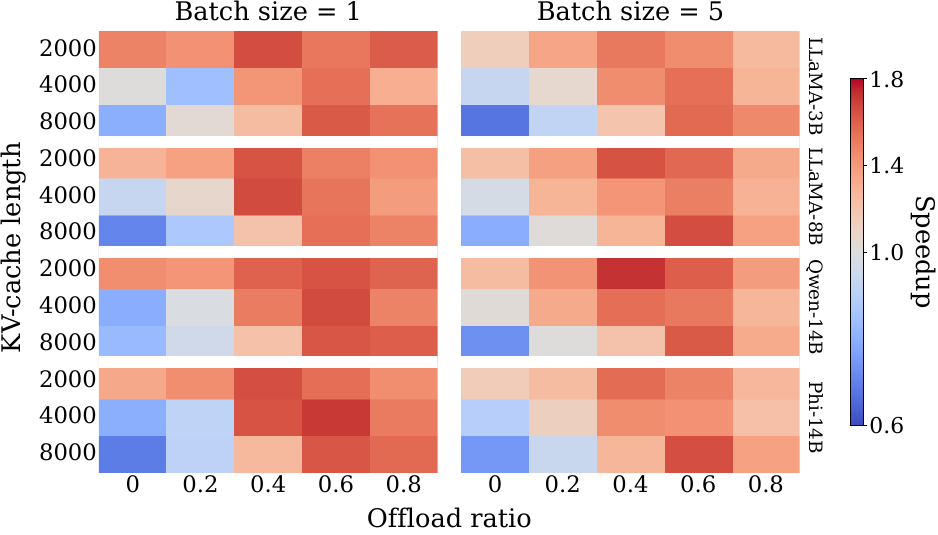}
    \vspace{-1em}
	\caption{Speedup over pure GPU attention offloading.}
	\label{fig:Speedup_over_GPU_based_attention}
    \vspace{-1.1em}
\end{figure}

\subsection{Decoding Performance}

\begin{figure*}[!t]
	\centering
	\includegraphics[width=1\textwidth]{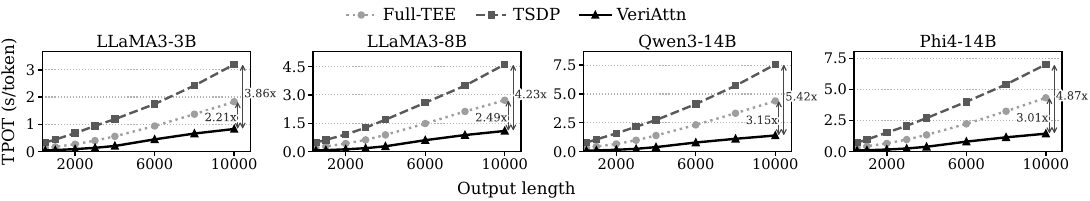}
    \vspace{-2em}
    \caption{Decoding TPOT comparison among Full-TEE, TSDP and \ctga across four models under varing output length.}
    \label{fig:full_model_decode_tpot}
    \vspace{-1em}
\end{figure*}
\begin{figure}[!t]
\centering
\begin{minipage}[t]{0.235\textwidth}
\vspace{0pt}
\centering
\includegraphics[
width=\linewidth,
height=2.2cm
]{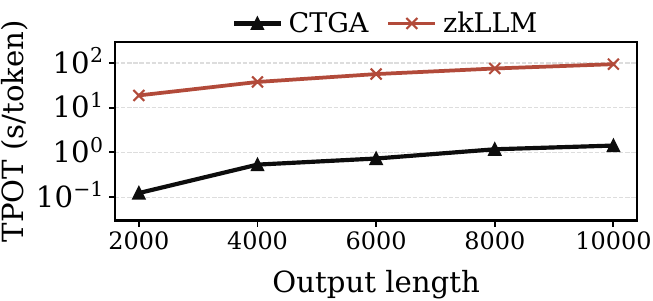}
\vspace{-2em}
\caption{\ctga vs. zkLLM in decoding TPOT on LLaMA3-8B.}
\label{fig:llama8b_decoding_attention_left}
\end{minipage}
\hfill
\begin{minipage}[t]{0.235\textwidth}
\vspace{0pt}
\centering
\includegraphics[
width=\linewidth,
height=2.2cm
]{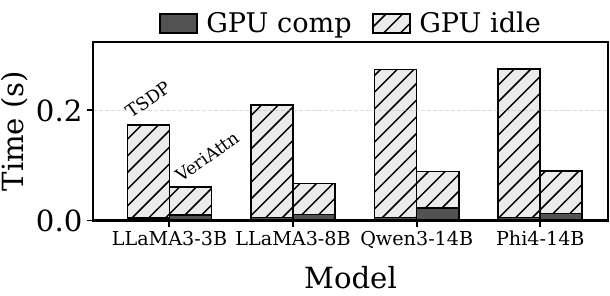}
\vspace{-2em}
\caption{GPU latency breakdown (decoding attention).}
\label{fig:decode-gpu-breakdown}
\end{minipage}
\vspace{-1em}
\end{figure}

\subsubsection*{Improvement over pure GPU offloading}
We evaluate the benefits of \ctga's collaborative TEE-GPU attention during decoding
across four LLMs, using pure GPU attention offloading (e.g., offload ratio $p=1$) as the baseline.
We set the GPU KV capacity to zero, so all KV entries are initially resident in TEE private memory.
Therefore, the pure GPU baseline must transfer all KV entries to GPU memory at each decoding step.
We sweep the KV cache size over 2,000, 4,000, and 8,000 tokens, 
and vary the offload ratio from $0$ to $1$ with a step size of $0.2$.
Figure~\ref{fig:Speedup_over_GPU_based_attention} shows the speedup achieved 
by collaborative TEE-GPU attention.
Across models, the best offload ratios are usually intermediate values, especially in the range from 0.4 to 0.6, and reach up to about $1.7\times$ speedup over $p=1$.
This suggests that \ctga benefits from keeping part of the attention computation inside the TEE
instead of transferring the entire KV cache to the GPU.
With an appropriate offload ratio, \ctga balances TEE-side computation, GPU acceleration, and the
cost of moving TEE-resident KV entries across the TEE-GPU boundary, making it effective for
long-context decoding.

In subsequent experiments, \ctga selects $p$ using a lightweight online profiling
heuristic. The scheduler monitors recent TEE/GPU computation time and TEE-GPU transfer time 
over a short window. If TEE computation repeatedly exceeds
the reusable idle budget, it increases $p$ to offload more non-resident KV entries
to the GPU. If the TEE remains idle while GPU execution is delayed by transfers,
it decreases $p$ so that more KV entries are processed locally inside the TEE.

\subsubsection*{TPOT in the decoding phase.}
Figure~\ref{fig:full_model_decode_tpot} compares decoding Time-Per-Output-Token (TPOT) for a
single request, i.e., batch size $1$, across Full-TEE, TSDP, and \ctga.
We use a fixed 500-token prompt and set the GPU KV-cache capacity to 2,000 tokens.
The figure reports average TPOT as output length increases.
The average TPOT is derived from measured per-token latency at different KV-cache lengths, capturing
the increasing decoding cost as the KV cache grows during generation.
\ctga consistently achieves the lowest TPOT across all models and output lengths.
Before the KV cache reaches the GPU capacity, most KV entries remain resident on the GPU, so the
decoding cost grows gradually.
After the KV cache exceeds the 2,000-token capacity, additional KV entries become non-resident and
long-context decoding becomes more expensive.
TSDP suffers from repeatedly moving non-resident KV-related attention states across the TEE-GPU
boundary, while Full-TEE performs the entire attention operation inside the TEE.
In contrast, \ctga uses collaborative attention execution to keep decoding
scalable under bounded GPU KV memory.
At an output length of 10,000, \ctga{} achieves TPOT speedups of $2.21$-$3.15\times$ over Full-TEE and $3.86$-$5.42\times$ over TSDP across the four models.
Figure~\ref{fig:llama8b_decoding_attention_left} further compares the TPOT latency of \ctga with the proof generation overhead of zkLLM on LLaMA3-8B.
zkLLM still incurs roughly two orders of magnitude higher latency than
\ctga during decoding.
These results show that \ctga reduces end-to-end decoding latency for long context
while operating within a fixed GPU KV-cache budget.

\begin{figure}[!t]
	\centering
	\includegraphics[width=0.46\textwidth,height=6.7cm]{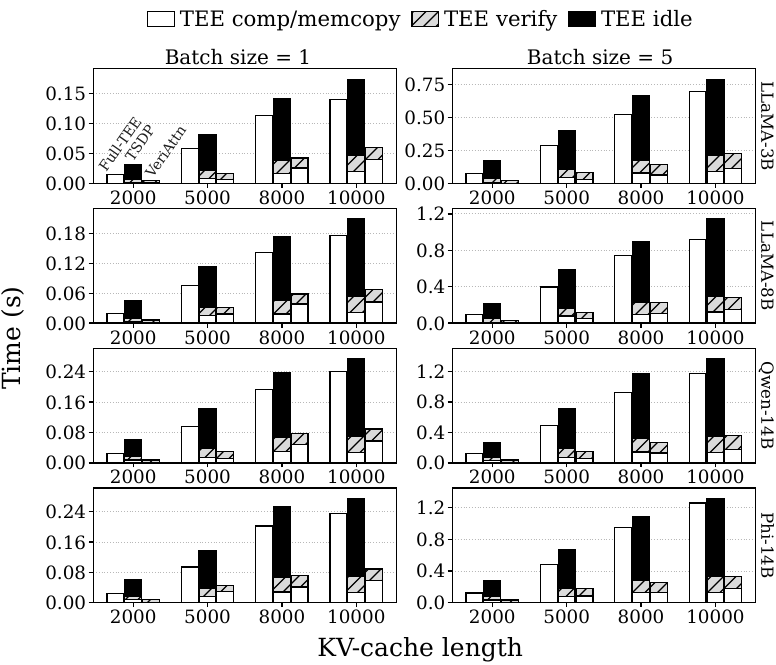}
    \vspace{-1em}
    \caption{
     Latency breakdown of decoding attention.
    }
    \label{fig:decoding-latency-Breakdown-ablation}
    \vspace{-1em}
\end{figure}

\subsubsection*{Decoding attention latency breakdown.}
Figure~\ref{fig:decoding-latency-Breakdown-ablation} isolates the decoding self-attention operator
and reports single-step attention latency at different KV-cache lengths.
We evaluate four models with KV-cache lengths from 2,000 to 10,000 tokens, batch sizes 1 and 5,
and a GPU KV-cache capacity of 2,000 tokens.
Each bar is decomposed into \textit{TEE comp/memcopy}, \textit{TEE verify}, and
\textit{TEE idle}, following the same definitions as in the prefill breakdown.
The breakdown explains the TPOT trend in Figure~\ref{fig:full_model_decode_tpot}.
Full-TEE is dominated by TEE-side attention computation, and its latency increases as the KV cache
grows.
TSDP offloads linear attention operations to the GPU, but keeps the non-linear
\texttt{SoftMax} path inside the TEE and follows a sequential TEE-GPU workflow.
Once the KV cache exceeds the GPU capacity, TSDP repeatedly transfers non-resident KV-related
intermediate states across the TEE-GPU boundary.
This creates substantial TEE idle time and limits the benefit of GPU offloading.
In contrast, \ctga offloads the GPU-side attention partition, including its
linear and non-linear operations, while processing the remaining TEE-resident KV
entries locally inside the TEE.
The GPU processes selected offloaded KV blocks, while the TEE locally processes the remaining
entries.
This collaborative execution reduces KV transfer volume and lowers TEE idle time during
long-context decoding.
At KV cache length 10,000 and batch size 1, \ctga is $2.31$-$2.70\times$ faster than Full-TEE and
$2.88$-$3.12\times$ faster than TSDP across the four models.
At batch size 5, \ctga is $3.11$-$3.86\times$ faster than Full-TEE and
$3.52$-$4.04\times$ faster than TSDP.
Figure~\ref{fig:decode-gpu-breakdown} also gives the complementary GPU-side view at KV cache length
10,000 and batch size 1.
Compared with TSDP, \ctga spends a larger fraction of time in useful GPU computation because it
offloads more of the attention path, including exponentiation and value aggregation to the GPU.
Meanwhile, \ctga substantially reduces GPU idle time across all four models, showing that its
KV partitioning and collaborative execution avoid the sequential transfer stalls that dominate TSDP.
These results show that \ctga improves decoding attention by reducing excessive
TEE-GPU transfers and using the compute resources of both the TEE and GPU.

\subsection{Verification Effectiveness and Output Quality}
\label{subsec:Verification-Effectiveness-and-Output-Quality}
We finally evaluate \ctga in terms of verification effectiveness and output quality.
For GPU offloading to be beneficial, 
TEE-side verification should be faster than direct TEE recomputation 
while still detecting corrupted results returned by the GPU.
For output quality, clean \ctga inference should preserve
standard full-attention inference within the calibrated numerical error budget.
All experiments in this subsection use WikiText validation prompts.

\subsubsection*{Verification efficiency and correctness.}
Table~\ref{tab:verification_effectiveness} 
reports verification cost and fault-injection results
on Qwen3-14B, where prefill uses a 6,000-token prompt and decoding uses a 10,000-token KV cache.
For each verification, 
we calibrate the numerical tolerance offline using clean FP16 GPU outputs.
The TEE computes FP32 verification residuals 
and sets the tolerance to twice the largest observed clean residual.
Each verification is run with 10 independent repetitions.
The verification checks are $19.5\times$ and $11.4\times$ 
faster than recomputation in prefill,
and $16.3\times$ and $13.8\times$ faster in decoding.

Next, we evaluate the verification correctness through fault injection that simulates the manipulations on GPU computations.
Each trial first runs attention normally and then corrupts GPU-returned tensor entries before TEE-side verification.
Each corrupted entry is sampled uniformly from the verified tensor and overwritten as
$
    x' = x + \alpha \cdot 10^{-2}\max(1, |x|),
$
where $\alpha\in\{-1,+1\}$ is chosen uniformly at random.
Table~\ref{tab:verification_effectiveness} 
reports 1,000 injected-fault trials and 1,000 clean
trials for each check.
\ctga detects all injected corruptions in both exponentiation and value aggregation, with no false rejections on clean executions.

\begin{table}[!t]
\centering
\caption{Verification cost and fault-injection results on Qwen3-14B.}
\label{tab:verification_effectiveness}
\vspace{-1em}
\setlength{\tabcolsep}{2.4pt}
\small
\begin{tabular}{llccccc}
\toprule
Phase & Check & Veri. & Comp. & Trials & Det. & False Rej. \\
\midrule
\multirow{2}{*}{Prefill}
  & \texttt{LPVeriExp}     & 0.12 s & 2.34 s & 1000 & 100.0\% & 0.0\% \\
  & \texttt{GVFAVeriMM}    & 0.036 s  & 0.41 s  & 1000 & 100.0\% & 0.0\% \\
\midrule
\multirow{2}{*}{Decode}
  & \texttt{LPVeriExp}     & 0.45 ms & 7.34 ms & 1000 & 100.0\% & 0.0\% \\
  & \texttt{GVFAVeriMM}    & 0.27 ms & 3.73 ms & 1000 & 100.0\% & 0.0\% \\
\bottomrule
\multicolumn{7}{p{\columnwidth}}{\footnotesize
\textit{Veri.} denotes verification latency using \texttt{LPVeriExp} for exponentiation and
\texttt{GVFAVeriMM} for value aggregation.
\textit{Comp.} denotes direct TEE
recomputation latency for the same operation. \textit{Det.} denotes the fraction of corrupted executions 
rejected by the verifier.
\textit{False Rej.} denotes the fraction of clean executions falsely rejected.
}
\end{tabular}
\vspace{-1.1em}
\end{table}

\begin{table}[!t]
\centering
\caption{Output quality of \ctga vs. standard full-attention inference across four models.}
\label{tab:output_quality}
\vspace{-1em}
\setlength{\tabcolsep}{2pt}
\fontsize{9.5pt}{11pt}\selectfont
\renewcommand{\arraystretch}{1.3}
\begin{tabular}{l|cc|c|cc}
\hline \rule{0pt}{12pt}
Model  & Max $|\Delta\mathrm{logit}|$ & Acc.
& Model  & Max $|\Delta\mathrm{logit}|$  & Acc. \\ [2pt]
\hline
LLaMA-3B  & $1.9{\times}10^{-3}$ & 1.00
& Qwen-14B & $3.1{\times}10^{-3}$ & 1.00 \\
LLaMA-8B  & $2.4{\times}10^{-3}$ & 1.00
& Phi-14B  & $2.8{\times}10^{-3}$ & 1.00 \\
\hline
\multicolumn{6}{p{\columnwidth}}{\footnotesize
\textit{Acc.} denotes ratio of evaluated prediction
positions whose top-1 token matches the reference. Max $|\Delta\mathrm{logit}|$ denotes
largest absolute logit difference.
}
\end{tabular}
\vspace{-1.1em}
\end{table}

\subsubsection*{Output quality.}
We evaluate output quality using deterministic greedy decoding to remove sampling randomness.
\ctga accepts attention results whose verification residuals fall within the calibrated numerical error budget.
We therefore examine whether such tolerated numerical variation affects final decoding decisions.
With each model generating 100 tokens from a 6,000-token prompt, we compare \ctga against standard
full attention inference.
Table~\ref{tab:output_quality} reports the ratio of evaluated prediction
positions whose top-1 token matches the reference and the
largest absolute logit difference.
\ctga achieves 100\% top-1 agreement across all tested models, with maximum logit differences below
$3.1\times10^{-3}$.
These results show that \ctga preserves full attention inference semantics within the proposed TEE-GPU attention framework in our evaluation.

%% file: Sections/Discussion.tex
\section{Discussion}
\label{sec:discuss}

\subsubsection*{Model and input confidentiality.}
This work focuses on computation integrity and communication efficiency. 
Model and input confidentiality can be complementary concerns in
practical deployments.
Some existing TEE offloading systems protect model weights and user inputs by keeping
secrets, such as masks, random transformations, or encryption keys, inside the
TEE, while offloading only protected representations to the untrusted
GPU~\cite{tramerslalom,sun2023shadownet,zhang2024no,wang2025game}.
These schemes are mainly designed for single matrix multiplications in linear layers.
Extending such protection to \ctga is non-trivial. 
To reduce communication,
\ctga does not require the GPU to return results after every attention
sub-operation. Instead, the GPU executes a chain of dependent linear and
non-linear operations and returns only selected intermediate states for TEE-side
processing and verification. Thus, the TEE must recover composed
attention states rather than the output of a single linear operator. Existing
mechanisms do not directly support such multi-step recovery, especially across
non-linear operations. The joint design of model/input confidentiality and communication efficient verifiable TEE-GPU attention is still a challenge for future research.

\subsubsection*{Support for online serving schedulers.} 
\ctga is compatible with existing schedulers for online LLM serving as an
attention backend within each worker. Modern serving systems improve throughput
under dynamic request streams through batching and scheduling at fine
granularity, including iteration-level scheduling and continuous
batching~\cite{yu2022orca,kwon2023efficient}, chunked prefill and hybrid
prefill-decode batching~\cite{agrawal2024taming}, and prefill-decode
disaggregation~\cite{zhong2024distserve,patel2024splitwise}. These schedulers
operate above the attention backend: they decide request admission, batch
formation, routing, and stage placement. Once a prefill chunk or decoding batch
is assigned to a worker equipped with a TEE and GPU, \ctga executes the
corresponding prefill or decoding workflow. This separation allows existing
schedulers to improve serving throughput at the system level, while \ctga
reduces TEE-GPU overhead within each assigned attention workload.

%% file: Sections/Conclusion.tex
\section{Conclusion}
\label{sec:conclude}

This paper presented \ctga, a communication-efficient TEE-GPU attention framework for
verifiable LLM inference. We showed that directly applying TSDP-style
partitioning to Transformer attention is inefficient: prefill incurs
quadratic-size attention-state transfers and expensive TEE-side
\texttt{SoftMax} computation, while long-context decoding repeatedly moves
non-resident KV-cache entries across the TEE-GPU boundary. \ctga addresses
these bottlenecks by offloading the main linear and non-linear attention
computation to the GPU, verifying GPU-returned attention states inside the TEE,
pipelining prefill execution, and partitioning decoding attention across
TEE- and GPU-resident KV blocks. Our evaluation shows that \ctga substantially
improves both full-model inference latency and attention-level performance over
existing computation integrity solutions, while preserving full attention
semantics under the evaluated settings.

%% file: Sections/Appendix.tex
\appendix
\newpage
\section{Appendix}

\subsection{Detailed Verification Procedures}
\label{app:Detailed-Verification-Procedures}

This appendix provides the detailed pseudocode for the two verification
procedures: \texttt{GVFAVeriMM} and \texttt{LPVeriExp} used by \ctga. The main text describes their verification principles,
while this appendix specifies the concrete offline randomness generation and
online verification steps. In both procedures, the TEE generates secret random
challenges during the offline phase and keeps them hidden from the GPU. During
online verification, the TEE checks the GPU-returned results using these secret
challenges and rejects immediately if any check fails.



\begin{algorithm}[!b]
\footnotesize
\caption{\texttt{GVFAVeriMM}: GVFA-based Verification for Value Aggregation}
\label{alg:gvfa-verimm}
\begin{algorithmic}[1]
\Require Mode $m\in\{\mathsf{prefill},\mathsf{decode}\}$, tolerance $\epsilon$, repetitions $\tau$, For prefill: $\mathbf E$, $\mathbf V$, claimed output $\widehat{\mathbf U}$, For decoding step $t$: $\mathbf e_t$, $\mathbf v_t$, claimed output $\widehat{\mathbf u}_t$, cached projections $\{\mathbf g_{t-1}^{(j)}\}_{j=1}^{\tau}$
\Ensure \textsc{Accept} or \textsc{Reject}

\Statex \textcolor{orange}{\textbf{Offline phase in the TEE.}}
\For{$j=1$ to $\tau$}
    \State Sample Gaussian vector
    $
        \mathbf w_G^{(j)} \sim \mathcal N(0,I_{d_h})
    $
    \State Initialize decoding projection cache
    $
        \mathbf g_0^{(j)} \gets [\,]
    $
\EndFor

\Statex \textcolor{orange}{\textbf{Online verification phase.}}

\If{$m=\mathsf{prefill}$}
    \For{$j=1$ to $\tau$}
        \State Compute projected value vector
        $
            \mathbf \eta^{(j)}
            \gets
            \mathbf V\mathbf w_G^{(j)}
        $
        \State Store
        $
            \mathbf g_L^{(j)} \gets \mathbf \eta^{(j)}
        $
        for later decoding
        \State Compute residual
        $
            \Delta
            \gets
            \left\|
            \mathbf E\mathbf \eta^{(j)}
            -
            \widehat{\mathbf U}\mathbf w_G^{(j)}
            \right\|_{\infty}
        $
        \If{$\Delta > \epsilon$}
            \State \Return \textsc{Reject}
        \EndIf
    \EndFor

\ElsIf{$m=\mathsf{decode}$}
    \For{$j=1$ to $\tau$}
        \State Update projection cache
        $
        \mathbf g_t^{(j)}
        \gets
        \begin{bmatrix}
        \mathbf g_{t-1}^{(j)} \\
        \mathbf v_t^\top \mathbf w_G^{(j)}
        \end{bmatrix}
        $
        \State Compute residual
        $
        \Delta
        \gets
        \big|
        \mathbf e_t \mathbf g_t^{(j)}
        -
        \widehat{\mathbf u}_t \mathbf w_G^{(j)}
        \big|
        $
        \If{$\Delta > \epsilon$}
            \State \Return \textsc{Reject}
        \EndIf
    \EndFor
\EndIf

\State \Return \textsc{Accept}

\end{algorithmic}
\end{algorithm}

\begin{algorithm}[!t]
\footnotesize
\caption{\texttt{LPVeriExp}: Log-Product Verification for Exponentiation}
\label{alg:lp-veriexp}
\begin{algorithmic}[1]
\Require Maximum sequence length $L_{\max}$, coefficient domain size $N_a$, Repetitions $\tau_a$, Gaussian checks $\tau_g$, tolerance $\epsilon'$, Rows to verify $\mathcal R$, valid index set $\mathcal I_r$ for each row $r$, Scores $\{s_{r,i}\}$, claimed exponentiation outputs $\{\widehat y_{r,i}\}$
\Ensure \textsc{Accept} or \textsc{Reject}

\Statex \textcolor{orange}{\textbf{Offline phase in the TEE.}}
\For{each possible row $r\le L_{\max}$}
    \For{$\ell=1$ to $\tau_a$}
        \State Sample secret coefficients
        $
            a_{r,i}^{(\ell)}
            \leftarrow
            \{1,\ldots,N_a\},
            i \in \mathcal I_r
        $
        \For{$k=1$ to $\tau_g$}
            \State Sample Gaussian scalar
            $
                w_{r,k}'^{(\ell)} \sim \mathcal N(0,1)
            $
        \EndFor
    \EndFor
\EndFor

\Statex \textcolor{orange}{\textbf{Online verification phase.}}

\For{each row $r\in\mathcal R$}

    \For{$\ell=1$ to $\tau_a$}
        \State Compute
        $
        R_r^{(\ell)} \gets \log\big(
        \prod_{i\in\mathcal I_r}
        \widehat y_{r,i}^{a_{r,i}^{(\ell)}}
        \big) - \sum_{i\in\mathcal I_r} a_{r,i}^{(\ell)} s_{r,i}
        $

        \For{$k=1$ to $\tau_g$}
            \If{$|R_r^{(\ell)}  w_{r,k}'^{(\ell)}| \ge \epsilon'$}
                \State \Return \textsc{Reject}
            \EndIf
        \EndFor
    \EndFor
\EndFor

\State \Return \textsc{Accept}
\end{algorithmic}
\end{algorithm}

\subsection{Proof of Lemma~\ref{lemma:lpverifyexp}}
\label{app:proof-nonlinear-verification}

We prove the false acceptance probability of \texttt{LPVeriExp}.
The TEE keeps the verification coefficients secret and checks whether
the exponentiation results returned by the GPU satisfy a log product
consistency relation.
A malicious GPU can pass the check only by constructing a forgery that is
consistent with the secret coefficients or by relying on the numerical
tolerance to hide a nonzero residual.

Consider the $r$-th row of the attention score matrix. Let
$
s_{r,i}
=
\mathbf q_r^\top \mathbf k_i / \sqrt{d_h}
+
\mathbf M_{r,i},
i\in\mathcal I_r,
$
where $\mathcal I_r$ denotes the valid, unmasked positions in row $r$.
The exact exponentiation output is
$
y_{r,i}=e^{s_{r,i}}.
$
In floating point execution, let $\bar y_{r,i}$ denote the value returned
by an honest GPU execution. The vector
$
\bar{\mathbf y}_r=[\bar y_{r,i}]_{i\in\mathcal I_r}
$
is considered valid if its honest residual satisfies the calibrated
tolerance. Specifically, define
\begin{equation}
R_r^{\mathrm{hon}}(\mathbf a)
=
\log\big(
\prod_{i\in\mathcal I_r}
\bar y_{r,i}^{a_i}
\big)
-
\sum_{i\in\mathcal I_r}a_i s_{r,i}.
\end{equation}
The tolerance $\epsilon'$ is calibrated so that honest floating point
executions satisfy
$
|R_r^{\mathrm{hon}}(\mathbf a)w'_{G}|<\epsilon'
$ for any $w'_{G}$.
Thus, the tolerance accounts for normal floating point rounding error in
honest GPU execution.

The GPU returns a claimed vector
$
\widehat{\mathbf y}_r
=
[\widehat y_{r,i}]_{i\in\mathcal I_r}.
$
If any returned value is not positive, NaN, or Inf, the TEE rejects
immediately. Otherwise, for each coefficient repetition
$\ell=1,\ldots,\tau_a$, the TEE samples a fresh secret coefficient vector
$
\mathbf a^{(\ell)}=[a_i^{(\ell)}]_{i\in\mathcal I_r},
a_i^{(\ell)}\leftarrow\{1,\ldots,N_a\},
$
where $\{1,\ldots,N_a\}$ is a nonzero integer coefficient domain of size
$N_a$. For each coefficient vector, the TEE also samples $\tau_g$
independent Gaussian scalars
$
w'_{G,\ell,1},\ldots,w'_{G,\ell,\tau_g}
\leftarrow
\mathcal N(0,1).
$
All coefficient vectors and Gaussian scalars are sampled inside the TEE
and are not revealed to the GPU before verification.

For exact exponentiation outputs, the following identity holds:
\begin{equation}
\log\big(
\prod_{i\in\mathcal I_r}
(e^{s_{r,i}})^{a_i^{(\ell)}}
\big)
=
\log\big(
e^{\sum_{i\in\mathcal I_r}a_i^{(\ell)}s_{r,i}}
\big)                                                    
=
\sum_{i\in\mathcal I_r}a_i^{(\ell)}s_{r,i}.
\end{equation}
Thus, \texttt{LPVeriExp} computes the residual
\begin{equation}
R_r(\mathbf a^{(\ell)})
=
\log\big(
\prod_{i\in\mathcal I_r}
\widehat y_{r,i}^{a_i^{(\ell)}}
\big)
-
\sum_{i\in\mathcal I_r}a_i^{(\ell)} s_{r,i}.
\end{equation}
The row is accepted only if all checks pass:
\begin{equation}
|R_r(\mathbf a^{(\ell)})w'_{G,\ell,j}|<\epsilon',
\quad
\ell=1,\ldots,\tau_a,
\quad
j=1,\ldots,\tau_g.
\end{equation}

For a possibly corrupted output $\widehat{\mathbf y}_r$, define the
logarithmic deviation as
$
\Delta_{r,i}
=
\log \widehat y_{r,i}
-
\log \bar y_{r,i}.
$
Let
$
T_r=\{i\in\mathcal I_r:\Delta_{r,i}\neq0\}
$
be the set of entries modified relative to the honest floating point
output, and let
$
n_{\mathrm{err},r}=|T_r|.
$
If $n_{\mathrm{err},r}=0$, the returned exponentiation results are
identical to the honest floating point outputs for row $r$.

Let 
\begin{equation}
\label{eq:log-product}
R_r^{\mathrm{adv}}(\mathbf a^{(\ell)})     =
    \log\big(
        \prod_{i\in\mathcal I_r}
        \widehat y_{r,i}^{a_i^{(\ell)}}
    \big)
    -
    \sum_{i\in\mathcal I_r}a_i^{(\ell)} s_{r,i}.
 \end{equation}
be the residual produced by
the possibly corrupted output $\widehat{\mathbf y}_r$. The additional
residual introduced by the attack is
\begin{equation}
\begin{aligned}
\widetilde R_r(\mathbf a^{(\ell)})
&=
R_r^{\mathrm{adv}}(\mathbf a^{(\ell)})
-
R_r^{\mathrm{hon}}(\mathbf a^{(\ell)})                    \\
&=
\sum_{i\in\mathcal I_r}
a_i^{(\ell)}
\big(
\log \widehat y_{r,i}
-
\log \bar y_{r,i}
\big)                                                     \\
&=
\sum_{i\in T_r}a_i^{(\ell)}\Delta_{r,i}.
\end{aligned}
\end{equation}
Therefore, the honest floating point residual is handled by the calibrated
tolerance, while $\widetilde R_r(\mathbf a^{(\ell)})$ captures the
additional residual caused by malicious modification.

\subsubsection*{Specific attack types.}
We first show that common output manipulations all reduce to the same
residual form
\begin{equation}
\widetilde R_r(\mathbf a^{(\ell)})
=
\sum_{i\in T_r}a_i^{(\ell)}\Delta_{r,i}.
\end{equation}

$\blacksquare$ {\bf Exponent domain tampering.}
The attacker may return
\begin{equation}
\widehat y_{r,i}
=
\bar y_{r,i}e^{\delta_i},
\qquad
\delta_i\neq0.
\end{equation}
Then
$
\Delta_{r,i}=\delta_i,
$
and the residual introduced by the attack becomes
\begin{equation}
\widetilde R_r(\mathbf a^{(\ell)})
=
\sum_{i\in T_r}a_i^{(\ell)}\delta_i.
\end{equation}

$\blacksquare$ {\bf Output domain additive tampering.}
The attacker may return
\begin{equation}
\widehat y_{r,i}
=
\bar y_{r,i}+\eta_i.
\end{equation}
If $\widehat y_{r,i}\le0$, the TEE rejects immediately. Otherwise,
\begin{equation}
\begin{aligned}
\Delta_{r,i}
&=
\log(\bar y_{r,i}+\eta_i)-\log \bar y_{r,i}                 \\
&=
\log\big(1+\frac{\eta_i}{\bar y_{r,i}}\big).
\end{aligned}
\end{equation}
Thus, additive tampering is also converted into a logarithmic error
$\Delta_{r,i}$.

$\blacksquare$ {\bf Position swapping or replay.}
The attacker may permute outputs or replay stale outputs. If
$
\widehat y_{r,i}=\bar y_{r,\pi(i)}
$
for a nonidentity permutation $\pi$, then
\begin{equation}
\Delta_{r,i}
=
\log \bar y_{r,\pi(i)}
-
\log \bar y_{r,i}.
\end{equation}
If the attacker replays a wrong value $y'_i$, then
\begin{equation}
\Delta_{r,i}
=
\log y'_i
-
\log \bar y_{r,i}.
\end{equation}
If the swapped or replayed values are identical to the honest floating
point values, there is no effective tampering. Otherwise, the attack
again induces a nonzero logarithmic error vector and is captured by
$
\widetilde R_r(\mathbf a^{(\ell)})
=
\sum_{i\in T_r}a_i^{(\ell)}\Delta_{r,i}.
$

Therefore, regardless of the concrete manipulation, an incorrect accepted
row must fall into one of two cases for each coefficient repetition:
either the attacker makes the residual introduced by the attack cancel
exactly, or a remaining nonzero residual passes the Gaussian tolerance
checks.

\subsubsection*{Forgery under secret coefficients.}
First, consider the case where the attacker tries to make the residual
introduced by the attack cancel exactly for the $\ell$-th coefficient
vector:
$
\widetilde R_r(\mathbf a^{(\ell)})
=
\sum_{i\in T_r}a_i^{(\ell)}\Delta_{r,i}
=
0.
$
If this exact cancellation succeeds, the corrupted output produces the
same log product residual as the honest floating point output under
$\mathbf a^{(\ell)}$ and is therefore accepted for this coefficient
repetition whenever the honest residual is accepted by the calibrated
tolerance.

If only one entry is modified, i.e., $T_r=\{j\}$, then
$
\widetilde R_r(\mathbf a^{(\ell)})
=
a_j^{(\ell)}\Delta_{r,j}.
$
Since $a_j^{(\ell)}\neq0$ and $\Delta_{r,j}\neq0$, we have
$
\widetilde R_r(\mathbf a^{(\ell)})\neq0.
$
Thus, a single modified exponentiation output cannot create exact
cancellation.
If $n_{\mathrm{err},r}\ge2$, the attacker may try to modify multiple
entries in a coordinated way so that their weighted logarithmic errors
cancel out:
$
\sum_{i\in T_r}a_i^{(\ell)}\Delta_{r,i}=0.
$
Since the GPU must commit to $\widehat{\mathbf y}_r$ before the TEE
samples $\mathbf a^{(\ell)}$, the logarithmic error vector
$\Delta_r=[\Delta_{r,i}]_{i\in T_r}$ is fixed independently of the secret
coefficients. Exact cancellation therefore requires the random
coefficient vector to fall into the cancellation hyperplane defined by
this fixed error vector.

Choose any index $j\in T_r$ with $\Delta_{r,j}\neq0$. For any fixed
assignment of the coefficients
$\{a_i^{(\ell)}: i\in T_r, i\neq j\}$, there is at most one real value of
$a_j^{(\ell)}$ that satisfies the cancellation equation:
$
a_j^{(\ell)}
=
-
\frac{
\sum_{i\in T_r, i\neq j}
a_i^{(\ell)}\Delta_{r,i}
}{
\Delta_{r,j}
}.
$
Since $a_j^{(\ell)}$ is sampled uniformly from
$\{1,\ldots,N_a\}$, the probability that the sampled value equals this
specific value is at most $1/N_a$. Therefore, for one fresh coefficient
vector,
\begin{equation}
\label{eq:exact-cancellation-bound}
\begin{aligned}
\Pr\big[
\widetilde R_r(\mathbf a^{(\ell)})=0
\big]
&\le
\mathbf 1\{n_{\mathrm{err},r}\ge2\}
\cdot
\frac{1}{N_a}.
\end{aligned}
\end{equation}

\subsubsection*{Gaussian tolerance bypass.}
If the attacker does not create exact cancellation for the $\ell$-th
coefficient vector, then the residual introduced by the attack remains
nonzero:
$
\widetilde R_r(\mathbf a^{(\ell)})\neq0.
$
The verifier checks the total residual
$
R_r^{\mathrm{adv}}(\mathbf a^{(\ell)})
$
rather than the attack component alone. Perturbations whose total
verification residual remains within the calibrated numerical error
budget are treated as floating point noise and are not considered
detectable attacks. For the remaining detectable perturbations, let
$\sigma'>0$ be a lower bound on the magnitude of the total verification
residual:
$
|R_r^{\mathrm{adv}}(\mathbf a^{(\ell)})|\ge\sigma',
$
whenever the corrupted output does not create exact cancellation and its
total residual lies outside the calibrated numerical error budget.

Conditioned on a fixed nonzero total residual,
we have $
R_r^{\mathrm{adv}}(\mathbf a^{(\ell)})w'_G
\sim
\mathcal N\big(0,R_r^{\mathrm{adv}}(\mathbf a^{(\ell)})^2\big).
$
Then, for one Gaussian repetition,
\begin{align}
&\Pr\big[
|R_r^{\mathrm{adv}}(\mathbf a^{(\ell)})w'_G|<\epsilon'
\mid |R_r^{\mathrm{adv}}(\mathbf a^{(\ell)})|\ge\sigma'
\big]\nonumber\\
& =
\Pr\big[
|w'_G|<
\frac{\epsilon'}{|R_r^{\mathrm{adv}}(\mathbf a^{(\ell)})|}
\big]                                                   
\le
\Pr\big[
|w'_G|<
\frac{\epsilon'}{\sigma'}
\big]                                                       \\
&=
2\Phi\big(\frac{\epsilon'}{\sigma'}\big)-1, \nonumber
\end{align}
where $\Phi(\cdot)$ is the cumulative distribution function of the
standard normal distribution.
Because the $\tau_g$ Gaussian scalars are independent, the probability
that all tolerance checks hide a nonzero total residual for the
$\ell$-th coefficient vector is bounded by 
$
\big(
2\Phi\big(\frac{\epsilon'}{\sigma'}\big)-1
\big)^{\tau_g}.
$

\subsubsection*{False acceptance probability for one row.}
Let $\mathsf{FA}_r$ denote the event that row $r$ contains at least one
maliciously modified exponentiation output but is still accepted.
Combining the exact cancellation bound
and the Gaussian tolerance bound,
one coefficient repetition accepts a corrupted row with probability at
most
$
\mathbf 1\{n_{\mathrm{err},r}\ge2\}
\cdot
\frac{1}{N_a}
+
\big(2\Phi(\frac{\epsilon'}{\sigma'})-1\big)^{\tau_g}.
$
With the $\tau_a$ coefficient repetitions,
we have
\begin{equation}
\begin{aligned}
\Pr[\mathsf{FA}_r]
\le
\left[
\mathbf 1 \{n_{\mathrm{err},r}\ge2 \}
\cdot
\frac{1}{N_a}
+
\big(
2\Phi\big(\frac{\epsilon'}{\sigma'}\big)-1
\big)^{\tau_g}
\right]^{\tau_a}.
\end{aligned}
\end{equation}
Since $0<2\Phi\big(\frac{\epsilon'}{\sigma'}\big)-1<1$, 
the false acceptance probability decreases with larger $N_a$, $\tau_a$, and $\tau_g$, 
and with a smaller ratio $\epsilon'/\sigma'$. 
Thus, it can be made small by appropriate parameter choices.

\subsubsection*{Numerical stability of \texttt{LPVeriExp}.}
\texttt{LPVeriExp} evaluates the log-product term in~\eqref{eq:log-product}.
Although the final quantity is logarithmic, directly forming
\begin{equation}
P_r^{(\ell)}
=
\prod_{i\in\mathcal I_r}
\widehat y_{r,i}^{a_i^{(\ell)}}
\end{equation}
and then computing $\log P_r^{(\ell)}$ is numerically unsafe. Over long
attention rows, the intermediate product can underflow to zero or overflow
to $\infty$ before the logarithm is applied. For example, finite positive
\texttt{float32} values cover only a bounded range, approximately
$[1.4\times 10^{-45}, 3.4\times 10^{38}]$. Once the product leaves this
range, the resulting logarithm no longer gives a valid finite residual.

To avoid this problem, the TEE does not store $P_r^{(\ell)}$ as an ordinary
floating-point product. It first rejects any returned value
$\widehat y_{r,i}$ that is non-positive, NaN, or Inf, and then accumulates
the product using a normalized representation
\begin{equation}
P_r^{(\ell)} = m\exp(c),
\qquad
\log P_r^{(\ell)} = \log m + c ,
\end{equation}
where the mantissa $m>0$ is kept in a safe range. Let $B>0$ be chosen with
sufficient margin for the working precision. After each elementary
multiplication used to form $\widehat y_{r,i}^{a_i^{(\ell)}}$ and the row
product, the verifier renormalizes
\begin{equation}
(m,c) \leftarrow
\begin{cases}
(m\exp(B),\, c-B), & m < \exp(-B),\\
(m\exp(-B),\, c+B), & m > \exp(B),\\
(m,c), & \text{otherwise}.
\end{cases}
\end{equation}
This update only transfers scale between $m$ and $c$ and therefore
preserves the represented log value:
\begin{equation}
\log(m\exp(\pm B)) + (c \mp B) = \log m + c .
\end{equation}

After all factors have been accumulated, \texttt{LPVeriExp} computes the
verification residual using the stable value $\log m+c$:
\begin{equation}
R_r^{\mathrm{adv}}(\mathbf a^{(\ell)})
=
(\log m+c)
-
\sum_{i\in\mathcal I_r} a_i^{(\ell)} s_{r,i}.
\end{equation}
The same Gaussian tolerance check is then applied to this residual. Thus,
the implementation realizes the original log-product verification relation
while avoiding intermediate underflow and overflow.

\begin{algorithm}[!t]
\footnotesize
\caption{\ctga Prefill Phase}
\label{alg:prefill_p}
\begin{algorithmic}[1]
\Require $Q,K,V$, number of heads $H$, sequence length $L$,
head block size $b_s$, exp row tile size $t_s$, causal mask $M$
\Ensure Prefill Attention Output $\mathbf{O}$
\Statex \textcolor{orange}{// TEE process}
\State $\mathcal{B}\gets\{1,\ldots,\lceil H/b_s\rceil\}$ \Comment{\texttt{HeadBlocks}}
\State $\mathcal{T}\gets\{1,\ldots,\lceil L/t_s\rceil\}$ \Comment{\texttt{RowTiles}}
\label{line:pre_init_RowTiles}

\ForAll{head block $b=[h_0,h_1)\in\mathcal{B}$}
    \State $\mathbf{Q}_{b},\mathbf{K}_{b},\mathbf{V}_{b}\gets \mathbf{Q}[:,h_0:h_1,:,:],\mathbf{K}[:,h_0:h_1,:,:],\mathbf{V}[:,h_0:h_1,:,:]$
    \label{line:pre_extract_block}
    \State \texttt{MemCpy} $(\mathbf{Q}_{b},\mathbf{K}_{b},\mathbf{V}_{b})$ from PRM to SHM
    \label{line:pre_copy_block}
\EndFor

\ForAll{head block $b\in\mathcal{B}$}
    \ForAll{row tile $s=[r_0,r_1)\in\mathcal{T}$}
        \State Wait until $\widehat{\mathbf{E}}_{b,s}$ is ready in SHM and \texttt{MemCpy} it to PRM
        \label{line:pre_recv_exp}
        \State $\mathbf{E}_{b,s} \gets$ \texttt{LPVeriExp}  $(\mathbf{Q}_{b},\mathbf{K}_{b},\widehat{\mathbf{E}}_{b,s})$ \textbf{if} verification passes 
        \State $\mathbf{Z}_{b,s}\gets \mathbf{E}_{b,s} \mathbf{1} $
        \label{line:pre_verify_exp}
        
    \EndFor
    \State $\mathbf{E}_{b} \gets\texttt{ConcatTiles}(\{\mathbf{E}_{b,s}\}_{s\in\mathcal{T}})$
    \State $\mathbf{Z}_{b}\gets\texttt{ConcatTiles}(\{\mathbf{Z}_{b,s}\}_{s\in\mathcal{T}})$\label{line:pre_concat_tiles}
    \State Wait until $\widehat{\mathbf{U}}_{b}$ is ready in SHM and \texttt{MemCpy} it to PRM
    \label{line:pre_recv_u}
    \State $\mathbf{U}_{b} \gets $\texttt{GVFAVeriMM}$(\mathbf{E}_{b},\mathbf{V}_{b},\widehat{\mathbf{U}}_{b})$ \textbf{if} verification passes
    \label{line:pre_verify_mm}
    \State $\mathbf{O}_{b}\gets \mathbf{U}_{b}\oslash \mathbf{Z}_{b}$ 
    \label{line:pre_norm} 
\EndFor
\State \Return $\mathbf{O}\gets\texttt{ConcatHeads}(\{\mathbf{O}_{b}\}_{b\in\mathcal{B}})$
\label{line:pre_return}

\Statex \textcolor{orange}{// GPU process }
\State Launch \texttt{PCIeXfer} the first ready $(\mathbf{Q}_{1},\mathbf{K}_{1},\mathbf{V}_{1})$ from SHM to GPU
\ForAll{head block $b\in\mathcal{B}$}
    \State Wait until $(\mathbf{Q}_{b},\mathbf{K}_{b},\mathbf{V}_{b})$ is ready on GPU
    \label{line:pre_gpu_copy_block}
    \State Launch \texttt{PCIeXfer} $(\mathbf{Q}_{b+1},\mathbf{K}_{b+1},\mathbf{V}_{b+1})$ from SHM to GPU once ready
    \State $\mathbf{S}_{b}\gets \mathbf{Q}_{b}\mathbf{K}_{b}^{\top}/\sqrt{d_h}+M_b$
    \label{line:pre_gpu_score}

    \ForAll{row tile $s=[r_0,r_1)\in\mathcal{T}$}
        \State $\widehat{\mathbf{E}}_{b,s}\gets\exp(\mathbf{S}_{b}[:,:,r_0:r_1,:])$
        \label{line:pre_gpu_exp}
        \State Launch \texttt{PCIeXfer} $\widehat{\mathbf{E}}_{b,s}$ from GPU to SHM
        \label{line:pre_gpu_return_exp}
    \EndFor

    \State $\widehat{\mathbf{E}}_{b}\gets\texttt{ConcatTiles}(\{\widehat{\mathbf{E}}_{b,s}\}_{s\in\mathcal{T}})$
    \label{line:pre_gpu_concat_exp}
    \State $\widehat{\mathbf{U}}_{b}\gets \widehat{\mathbf{E}}_{b}\mathbf{V}_{b}$
    \label{line:pre_gpu_uv}
    \State Launch \texttt{PCIeXfer} $\widehat{\mathbf{U}}_{b}$ from GPU to SHM
    \label{line:pre_gpu_return_u}
\EndFor
\end{algorithmic}
\end{algorithm}

\subsection{Detailed Prefill and Decoding Workflows}
\label{app:detailed_workflows}
This section gives detailed pseudocode for the prefill and decoding workflows of
\ctga. The main text focuses on the design principles, while this appendix
specifies the concrete data movement, GPU computation, and TEE-side
verification steps. 
Because VM-based TEEs, such as Intel TDX, isolate private
memory from untrusted devices, GPU communication must be staged through the
PRM-SHM-GPU data path. \ctga therefore implements each workflow with two
concurrent processes: a TEE process for PRM-SHM copies, local attention or
verification, and normalization, and a GPU process for SHM-GPU transfers and
GPU kernels. A single synchronous process would serialize these operations,
because waiting for a transfer or GPU result would prevent the TEE from
performing useful local computation or verification. The two-process design
realizes the overlap between TEE-side work, data movement, and GPU execution in
the prefill and decoding workflows.

In the algorithms, \texttt{MemCpy} denotes a TEE-side memory copy between PRM
and SHM, and \texttt{PCIeXfer} denotes a PCIe transfer between SHM and GPU
memory. All copy and kernel launches are asynchronous unless a \texttt{Wait}
statement explicitly enforces a data dependency. GPU-returned tensors are
written with hats before verification and without hats after the TEE accepts
them.

\subsubsection{Prefill Pseudocode}
\label{app:prefill-workflow}
Algorithm~\ref{alg:prefill_p} shows the detailed prefill workflow. 
It uses the same notation as Section~\ref{subsec:prefill_stage}. 
The TEE process
performs preprocessing by copying input head blocks from PRM to SHM, and
performs postprocessing by copying GPU results from SHM to PRM, verifying them,
accumulating row sums, and normalizing accepted outputs. The GPU process
transfers prepared blocks from SHM to GPU memory and executes attention
kernels. Head-block partitioning overlaps PRM-SHM memory copy, SHM-GPU
transfer, and GPU computation. Row-tile partitioning further overlaps
exponentiation transfer, TEE-side verification, and GPU computation.

\subsubsection{Decoding Pseudocode}
\label{app:decoding_workflow}

Algorithm~\ref{alg:decode_p} shows the detailed decoding workflow at step $t$.
The TEE maintains
the complete protected KV cache $\mathbf{KV}_{tee}$, while the GPU maintains a
bounded resident cache $\mathbf{KV}_{gpu}$. If $\mathbf{KV}_{gpu}$ has available
capacity, the new KV entry is appended to both caches. Otherwise, \ctga forms
the non-resident cache
$\mathbf{KV}_{tee}\setminus\mathbf{KV}_{gpu}$ and splits it
into $\mathbf{KV}_{off}$ and $\mathbf{KV}_{rem}$ using the offload ratio $p$.
The GPU processes $\mathbf{KV}_{gpu}$ and the selected offloaded blocks from
$\mathbf{KV}_{off}$, while the TEE locally processes $\mathbf{KV}_{rem}$. The
TEE verifies the 
GPU-returned exponentiation and value aggregation states and then
merges the verified GPU-side state with the TEE-side state.

\begin{algorithm}[!t]
{\fontsize{8pt}{9pt}\selectfont
\caption{\ctga Decoding Phase at step $t$}
\label{alg:decode_p}
\begin{algorithmic}[1]
\Require $q_t,k_t,v_t$, TEE KV cache $\mathbf{KV}_{tee}$, GPU KV cache $\mathbf{KV}_{gpu}$, 
GPU KV cache capacity $C_{gpu}$, offload ratio $p$, KV block size $b_s^{\prime}$
\Ensure Decoding Attention Output $\mathbf{O}$
\Statex \textcolor{orange}{// TEE process}
\State $\mathbf{KV}_{tee}\gets \mathbf{KV}_{tee}\Vert(\mathbf{k}_t,\mathbf{v}_t)$
\label{line:dec_append_tee}

\If{$|\mathbf{KV}_{gpu}|<C_{gpu}$}
    \State $\mathbf{KV}_{off},\mathbf{KV}_{rem},\mathcal{B}_{kv}\gets\emptyset,\emptyset,\emptyset$
    \State \texttt{MemCpy} $(q_t,k_t,v_t)$ from PRM to SHM
    \label{line:dec_copy_qkv}
\Else
    \State $\tau\gets\lfloor p\cdot |KV_{tee}\setminus KV_{gpu}|\rfloor$
    \State $KV_{off}\gets KV_{tee}\setminus KV_{gpu}[:,:,:\tau,:]$
    \State $KV_{rem}\gets KV_{tee}\setminus KV_{gpu}[:,:, \tau:,:]$ 
    \State $\mathcal{B}_{kv}=\{1,...,\lceil |KV_{off}|/b'_s\rceil\}$
       \Comment{\texttt{KVBlocks}} 
    \State \texttt{MemCpy} $q_t$ from PRM to SHM
    \label{line:dec_copy_q}
    \ForAll{$b\in\mathcal{B}_{kv}$}
        \State \texttt{MemCpy} $\mathbf{KV}_{off,b}$ from PRM to SHM
    \EndFor
\EndIf

\State $(\mathbf{U}_{tee},\mathbf{Z}_{tee})\gets\texttt{TEEAttnComp}(\mathbf{q}_t,\mathbf{KV}_{rem})$
\label{line:dec_local_sum}

\State Wait until $\widehat{\mathbf{E}}_{gpu}$ is ready in SHM and \texttt{MemCpy} it to PRM
\label{line:dec_recv_hot}
\State $\mathbf{E}_{gpu} \gets$\texttt{LPVeriExp}  $(\mathbf{q}_t,\mathbf{K}_{gpu},\widehat{\mathbf{E}}_{gpu})$ \textbf{if} verification passes
\State $\mathbf{Z}_{gpu}\gets\mathbf{E}_{gpu} \mathbf{1}$
\label{line:dec_verify_hot}

\ForAll{$b\in\mathcal{B}_{kv}$}
    \State Wait until $\mathbf{E}_{b}$ is ready in SHM and \texttt{MemCpy} it to PRM
    \label{line:dec_recv_eb}
    \State $\mathbf{E}_{b} \gets$ \texttt{LPVeriExp}  $(\mathbf{q}_t,\mathbf{K}_{off,b},\widehat{\mathbf{E}}_{b})$ \textbf{if} verification passes
    \State $\mathbf{Z}_{gpu}\gets \mathbf{Z}_{gpu}+\mathbf{E}_{b} \mathbf{1}$
\EndFor\label{line:dec_verify_off}

\State Wait until $\widehat{\mathbf{U}}_{gpu}$ is ready in SHM and \texttt{MemCpy} it to PRM
\label{line:dec_recv_u}
\State $\mathcal{E}\gets\{\mathbf{E}_{gpu}\}\cup\{\mathbf{E}_{b}\}_{b\in\mathcal{B}_{kv}}$, \quad
       $\mathcal{V}\gets\{\mathbf{V}_{gpu}\}\cup\{\mathbf{V}_{off,b}\}_{b\in\mathcal{B}_{kv}}$

\State $\mathbf{U}_{gpu} \gets$\texttt{GVFAVeriMM}$(\mathcal{E},\mathcal{V},\widehat{\mathbf{U}}_{gpu})$ \textbf{if} verification passes

\label{line:dec_verify_mm}
\State \Return $\mathbf{o}_t\gets (\mathbf{U}_{gpu}+\mathbf{U}_{tee})/(\mathbf{Z}_{gpu}+\mathbf{Z}_{tee})$
\label{line:dec_merge}

\Statex \textcolor{orange}{// GPU process}

\If{$|KV_{gpu}|<C_{gpu}$}
    \State Wait until $(\mathbf{q}_t, \mathbf{k}_t,\mathbf{v}_t)$ is ready in SHM and \texttt{PCIeXfer} them to GPU
    \State $\mathbf{KV}_{gpu}\gets \mathbf{KV}_{gpu}\Vert(\mathbf{k}_t,\mathbf{v}_t)$ once $(\mathbf{k}_t,\mathbf{v}_t)$ is ready in GPU
    \label{line:dec_gpu_append}
\EndIf

\State $\widehat{\mathbf{E}}_{gpu}\gets\exp(\mathbf{q}_t\mathbf{K}_{gpu}^{\top}/\sqrt{d_h})$
\label{line:dec_gpu_hot_exp}
\State Launch \texttt{PCIeXfer} $\widehat{\mathbf{E}}_{gpu}$ from GPU to SHM
\label{line:dec_gpu_hot_return}
\State $\widehat{\mathbf{U}}_{gpu}\gets \widehat{\mathbf{E}}_{gpu}\mathbf{V}_{gpu}$
\label{line:dec_gpu_hot_uv}

\State Wait until $\mathbf{q}_t, \mathbf{KV}_{off,1}$ is ready in SHM and \texttt{PCIeXfer} them to GPU
\ForAll{$b\in\mathcal{B}_{kv}$}
    \State Wait until $\mathbf{KV}_{off,b}$ is ready in GPU 
    \State Launch \texttt{PCIeXfer} $\mathbf{KV}_{off,b+1}$ from SHM to a temp. GPU workspace\label{line:dec_gpu_kvoff}
    \State $\widehat{\mathbf{E}}_{b}\gets\exp(\mathbf{q}_t \mathbf{K}_{off,b}^{\top}/\sqrt{d_h})$
    \label{line:dec_gpu_eb}
    \State Launch \texttt{PCIeXfer} $\widehat{\mathbf{E}}_{b}$ from GPU to SHM
    \label{line:dec_gpu_eb_return}
    \State $\widehat{\mathbf{U}}_{gpu}\gets \widehat{\mathbf{U}}_{gpu}+\widehat{\mathbf{E}}_{b} \mathbf{V}_{off,b}$
    \label{line:dec_gpu_accum}
\EndFor \label{line:dec_gpu_accum_end_for}

\State Launch \texttt{PCIeXfer} $\widehat{\mathbf{U}}_{gpu}$ from GPU to SHM
\label{line:dec_gpu_return_u}
\end{algorithmic}
}
\end{algorithm}